\documentclass[journal]{IEEEtran}
\usepackage{array}
\usepackage{graphicx} 
\usepackage{color} 
\usepackage{xcolor}
\usepackage{cite}
\usepackage{booktabs}
\usepackage{multirow}
\usepackage{amsmath}
\usepackage{amsfonts} 
\usepackage{amssymb} 
\usepackage{mathrsfs}
\usepackage{graphicx}
\usepackage{textcomp}
\usepackage{subfigure} 
\usepackage{bm}
\usepackage{threeparttable}

\usepackage{amsthm}

\usepackage{multirow}
\usepackage{subfigure}  
\usepackage{color}
\usepackage{makecell}
\usepackage{float}
\usepackage[ruled,linesnumbered]{algorithm2e}
\usepackage{algpseudocode}
\makeatletter
\newcommand{\removelatexerror}{\let\@latex@error\@gobble}
\makeatother

\begin{document}
\title{A Heterogeneous Graph Convolution based Method for Short-term OD Flow Completion and Prediction in a Metro System}
\author{Jiexia~Ye, ~Juanjuan~Zhao*,  ~Furong~Zheng, ~Chengzhong~Xu, IEEE~Fellow\\
\thanks{*Corresponding Author: Juanjuan Zhao}
\thanks{
 Jiexia Ye is with Guangdong-Hong Kong-Macao Joint Laboratory of Human-Machine Intelligence-Synergy Systems, Shenzhen Institute of Advanced Technology, Chinese Academy of Sciences, China (E-mail: \{zumri.jiexiaye\}@um.edu.mo).

Juanjuan Zhao, Furong Zheng are with Shenzhen Institutes of Advanced Technology, Chinese Academy of Sciences, China (E-mail: \{jj.zhao, fr.zheng\}@siat.ac.cn).

Chengzhong Xu is with State Key Lab of IOTSC, Department of Computer Science, University of Macau, Macau SAR, China (E-mail: czxu@um.edu.mo).
}
}

\maketitle
\begin{abstract}
Short-term OD flow (i.e. the number of passenger traveling between stations) prediction is crucial to traffic management in metro systems. Due to the delayed effect in latest complete OD flow collection, complex spatiotemporal correlations of OD flows in high dimension, it is more challengeable than other traffic prediction tasks of time series.
Existing methods need to be improved due to not fully utilizing the real-time passenger mobility data and not sufficiently modeling the implicit correlation of the mobility patterns between stations.
In this paper, we propose a Completion based Adaptive Heterogeneous Graph Convolution Spatiotemporal Predictor. The novelty is mainly reflected in two aspects.
The first is to model real-time mobility evolution by establishing the implicit correlation between observed OD flows and the prediction target OD flows in high dimension based on a key data-driven insight: the destination distributions of the passengers departing from a station are correlated with other stations sharing similar attributes (e.g. geographical location, region function).
The second is to complete the latest incomplete OD flows by estimating the destination distribution of unfinished trips through considering the real-time mobility evolution and the time cost between stations, which is the base of time series prediction and can improve the model's dynamic adaptability.
Extensive experiments on two real world metro datasets demonstrate the superiority of our model over other competitors with the biggest model performance improvement being nearly 4\%. In addition, the data complete framework we propose can be integrated into other models to improve their performance up to 2.1\%.
\end{abstract}

\section{Introduction}
\label{lab:Introduction}
\IEEEPARstart{R}ail transit plays a key role in the comprehensive transportation system and becomes a popular travel mode for passengers because of its safety, speed, punctuality, large traffic volume. In recent years, with the expansion of transportation network and the improvement of accessibility, more and more people choose to use rail transit for travel, and the large-scale passenger flow is becoming more and more normal. Accurate and real-time prediction of the spatiotemporal distribution of passenger flow is the premise of ensuring the safe and efficient operation of rail transit, and has attracted increasing interest.

Short-term passenger flow prediction in URT focuses on utilizing historical ridership information to predict the future passenger demand from few minutes to few hours. It can be divided into three specific tasks, i.e. entry/exit flow prediction, origin-destination (OD) flow prediction, and sectional flow prediction\cite{zhang2019novel}. To date,  extensive effort has been devoted to entry/exit demand prediction which aims to forecast the number of passengers entering or exiting each station \cite{lu2021dual, chen2020physical, li2020tensor, wei2012forecasting, ma2018parallel, liu2020physical, mulerikkal2022performance, wang2021metro, ou2020stp, zhang2020deep, zhang2020multi, Li2020TensorCF, liu2020physical}. By comparison, OD flow prediction which further forecasts the passenger destination distribution of each origin station
receives much less attention. Although the station-level entry/exit prediction is useful, it is still coarse and inadequate for manager to implement network-wise real-time monitoring and proactive operations.

The real-time passenger mobility information provided by OD prediction can better support metro systems for train scheduling, fine-grained abnormal flow warning, passenger route planning and it is also an essential input for sectional passenger flow prediction task. For example, the OD flow is the basis to get the travel demand of each metro line. The manager can design appropriate strategy through shortening or extending the train headway of corresponding metro line to meet the dynamic travel demand. In addition, the entry/exit prediction can only detect the abnormal ridership departing from or destining to each station, but unable to detect the fine-grained flows such as the passengers waiting at each platform of a station, and the transfer flow at each transferring station while OD demand prediction can provide information to detect such abnormality. Accurate OD demand can help manager to recommend proper routes for passengers to achieve global traffic balance in whole metro system.

Different from entry/exit prediction in URT which aims to predict a $N$ dimension vector of passenger flows departing or arriving at each station in a metro network with $N$ stations, OD flow prediction model outputs an OD matrix with a high dimensionality of $N^2$. That increases its uncertainties, and makes us pay more attention to various passenger mobility patterns. 
Passenger mobility is composed of both periodicity (7-day or 24-hour periodicity) and randomicity. The periodicity is mainly due to the constraint of people's working and living during the day and at night, which can be easily observed from the traffic AM and PM rush hours and different traffic patterns on weekdays and weekends. The randomicity may be caused by passengers's non-fixed activities (e.g. entertainment, shopping) or some external factors such as celebrations, exhibitions, weather. However in most cases, it is difficult to obtain the real-time travel plans of passengers or external factors. Therefore, how to model the randomicity is the biggest challenge of OD flow prediction. In general, the randomicity exhibits local or global mobility variability.

The local mobility variability occurs between two stations within a period of time. That may be caused by a celebration party organized by an enterprise, where the OD flow between the enterprise and celebration locations has a certain fluctuation. The local mobility evolution can be learned from latest OD flows due to the recent dependency as that of entry/exit flow. However, it's impractical to feed the OD flows at last time slots into the model because the finished OD flows (i.e. the OD flows collected before the predicted target time slot) are likely to be incomplete due to unfinished trips and the full OD flows can only be obtained after all the passengers finish their journeys. Such real-time data availability needs to be tackled to capture mobility dynamicity for accurate prediction.

The global flow fluctuation occurs between multiple stations over a period of time. It may be caused by some events (concert, athletic meeting, opening of shopping mall). The mobility fluctuations of different stations might be asynchronous and the mobility dynamic patterns can be learned from each other. For example, the opening of a shopping mall located at station D may lead to passenger mobility pattern changing in multiple surrounding stations (e.g. A, B, C). To predict the OD flow of the A-D pair, the mobility patterns of OD flows of B-D and C-D pairs can be beneficial. The establishment of the spatial correlations can help us to select related features to the prediction target from high-dimensional observation data, so as to model passenger mobility patterns of each station more effectively even with insufficient training data. Therefore, how to model the spatial correlations needs to be resolved.

To date, some conventional methods are adopted for OD matrix prediction by modeling the temporal correlation of time series, such as state space model \cite{yao2015real, chen2017short},  vector autoregression \cite{cheng2021real}, probabilistic model \cite{dai2018short}, clustering method \cite{lin2007generalized}, matrix decomposition \cite{gong2018network, gong2020online}.
These methods are applicable to the scenes with stable passenger flow. Due to its limited feature extraction and learning ability, it is not suitable for the OD flow prediction with nonlinear and complex spatiotemporal correlations.

Recently, some advanced deep learning architectures based on Long Short-Term Memory Network \cite{li2020spatio, toque2016forecasting, zhang2019short}, Convolutional Network \cite{zhang2021short, noursalehi2021dynamic, he2022short}, Graph Convolutional Network \cite{jiang2022deep, liu2022online} are proposed to learn the spatiotemporal flow dependency of stations to predict the OD flows of various traffic scenarios. However, most of these methods assume that the latest OD flows are available, or not sufficiently use the latest incomplete OD flows,  which are not applicable for metro scenes. Among them, to the best of our knowledge,
 \cite{zhang2021short} is the most related work with this research. It proposed CAS-CNN with channel-wise attention mechanism and split CNN for large OD value prediction. It introduced inflow/outflow-gated mechanism to solve the data availability challenge but it just considered the real-time inflows and outflows, totally abandoned the incomplete real-time OD matrixes which might loss accurate real-time mobility information. Besides, the biggest limitation is failing to exploit the global spatial dependency between the mobility patterns of stations.

In this paper, we propose a model called \textbf{C}ompletion based \textbf{A}daptive \textbf{H}eterogeneous \textbf{G}raph \textbf{C}onvolution \textbf{S}patiotemporal \textbf{P}redictor (C-AHGCSP) for metro OD prediction. It can be used both for recent OD flow completion and future OD flow prediction. The OD flow prediction is a typical problem of time series in high dimensions. We select some relevant features from latest OD flows for target OD flow prediction to avoid high-dimensional disaster based on the insight that: the destination distributions of the passengers
departing from a station are correlated with other stations sharing similar static or dynamic attributes. For example,
the groups of passengers distributed around geographically adjacent metro stations are more inclined to show similar mobility pattern than those distant ones. The metro stations in regions with similar region functions (entertain, education, business  and residence) are possible to have similar mobility patterns. In addition, the mobility patterns of the stations with similar flow changing trend in the latest period may also be similar in the future. These spatial correlations between stations with similar attributes (geographical location, region function, recent flow changing  variation) not only can provide mutual enhancement and complementation of mobility prediction for each other, but also help us to extract more effective features from high dimensional input to predict the target OD flows.

In addition, the time series of latest OD flows is the base of OD flow prediction. We convert latest OD flow completion issue into destination distribution estimation of the unfinished trips. Given a past time slot with incomplete OD flows, our idea is first to use the time series of complete OD flows before to predict the OD flows, so that the destination distribution of all trips (contain finished trips and unfinished trips) can be obtained. Then the destination distribution of unfinished trips can be estimated by considering time cost between stations.
Overall, the main contributions of our paper are as follows:
\begin{itemize}
\item {We proposes an innovative method C-AHGCSP both for latest OD flow completion and OD flow prediction, which can be used to capture the recent and future fine-grained passenger flow distributions on entire metro network. The OD completion method as a sub-module can also be integrated into  existing prediction models to improve their performance.}
\item{We propose a data complete estimator composed of a prior estimator and AHGCSP based estimator to estimate the full recent OD matrix sequence by taking advantage of the dynamic mobility patterns, unfinished trips, different time cost of OD pairs.}
\item {Based on the insight that the stations with similar attributes are possible to have similar mobility patterns, we design a module named AHGCSP to learn the global spatiotemporal correlations. That enables our model to extract effective features from high-dimensional input and improve model robustness even with limited  training data. It first organizes latest OD flows into a sequence of dynamic station graph with heterogeneous edges (geographical similarity by Gaussian Kernel, region function similarity by KL divergence, and real-time mobility similarity by self-attention mechanism). Then Graph convolution network based adaptive heterogeneous module and LSTM are used to learn the dynamic spatiotemporal mobility evolution trend. }
\item{Extensive experiments on large-scale real-world metro datasets demonstrate the superiority of our model, compared with baselines. We have published our code and one of our datasets in  https://github.com/start2020/C-AHGCSP.}

\end{itemize}

\section{Related Works}
Short-term OD flow (the number of passenger traveling between stations) prediction is crucial to traffic management. In recent years, with the increased availability of data, many researchers have begun to use the various methods to predict OD flows in various  traffic scenario, such as road/city OD matrix \cite{saberi2017complex, lin2007generalized, hu2017sequential, ma2013day, zhou2006dynamic, tang2021dynamic, xiong2020dynamic, ren2017efficient, ou2019learn, ye2016optimal, huang2019origin, he2018origin, ma2018statistical}, ride-hailing OD matrix \cite{zhang2021dneat, wang2019origin, shi2020predicting}, taxi OD matrix \cite{liu2019contextualized, chu2019deep, zhang2017deep, lim2021origin, duan2019prediction, yao2020spatial, hu2020stochastic, tong2017simpler, miao2021deep, ke2021predicting, wang2020multi}, as well as bus OD matrix \cite{jin2022spatio, wang2011bus, chen2011short,jafari2021procedure}.
However, the contextual information of the above tasks is different from that of metro system. First, the real-time data availability in these tasks is various and different. As mentioned in the last section, in metro scenario, the latest OD matrices is incomplete due to unfinished trips before the predicted time slot. However, in the case of ride-hailing services, the latest OD matrix is complete, which can be obtained directly when passengers call for car service in related apps. As to bus system (except the bus rapid transit (BRT)), the full OD matrices are totally unavailable even when all trips are finished due to that only boarding station is recorded without the alighting station. The cameras and sensors in road network can only detect section flow instead of the origins or destinations of vehicles. The availability of OD matrices in taxi OD task is similar to that in URT but the boarding and alighting points of taxi are not fixed and previous studies usually divided the whole research region into grid-based origin and destination zones, making it significantly different from subway systems. The different contextual information above makes it hard to transfer the OD prediction models in other traffic scenarios to metro scenario.

For metro scenario, some conventional methods are adopted for OD matrix prediction by modeling the temporal correlation of time series, such as state space model \cite{yao2015real, chen2017short}, vector autoregression \cite{cheng2021real}, probabilistic model \cite{dai2018short}, clustering method \cite{lin2007generalized} and so on \cite{zheng2019dynamic, yang2020dynamic}. However, these methods build a model for each OD pair, which may lead to inefficiency when the number of OD pairs is very large. In addition, they are only applicable to relatively stable passenger flow prediction due to the limited nonlinear correlation learning. Some researches use matrix decomposition or its variants to predict OD passenger flow\cite{gong2018network, gong2020online}. Compared with other conventional methods, matrix decomposition based methods considering all OD pairs as a whole can model the stable global correlation between OD pairs. However, these methods are linear techniques in nature, and not suitable for modeling the global nonlinear correlation of OD flows. In addition, they requires a complex decomposition calculation process, which limits the scalability when the number of stations is extremely large. In recent years, for metro scenario, some advanced deep learning architectures based on Long Short-Term Memory Network \cite{li2020spatio, toque2016forecasting, zhang2019short}, Convolutional Network \cite{zhang2021short, noursalehi2021dynamic, he2022short}, Graph Convolutional Network \cite{jiang2022deep, liu2022online} are proposed to solve the metro OD prediction problem. Noursalehi et.al \cite{noursalehi2021dynamic} proposed a scalable methodology for realtime OD demand prediction. It extracted the local spatial features by a channel-wise attention block with a squeeze-and-excitation layer. Liu et.al \cite{liu2022online} proposed a heterogeneous information aggregation machine to fully take advantage of historical data, e.g. incomplete OD matrices, unfinished order vectors, and DO matrices for OD and OD ridership multi-step forecasting. But they all ignored the time-evolving global passenger mobility pattern correlation in metro OD flow which is valuable for improving the prediction accuracy. To the best of our knowledge, \cite{zhang2021short} is the most related work with this research. It proposed CAS-CNN with channel-wise attention mechanism and split CNN for large OD value prediction. It introduced inflow/outflow-gated mechanism to solve the data availability challenge but it just considered the real-time inflow and outflow information and totally abandoned the incomplete real-time OD matrix which might loss accurate real-time destination distribution of passengers. Besides, the method relies on multiple CNN layers to automatically learn the global spatial correlation between different OD pairs. The method may be able to learn the global correlation if sufficient training data is available, but not suitable for OD predictions in the scenarios with limited data source (for instance one month) and the dimension of OD pairs is very large. This paper aims to make full use of dynamic observation data and complex factors influencing passenger flow mobility between stations to model the global correlation.

\begin{table*}[h]
\caption{The definition of terminologies in this paper}
\begin{tabular} {p{110pt}|p{150pt}|p{220pt}}
\toprule
\textbf{Terminology} &\textbf{Notation}  &\textbf{Description}\\ 
\hline
\textbf{Time Slot} & $t/t'$  &  Each day is divided evenly into $\text{T}$ time slots with a time granularity $\delta$. t represents an arbitrary time slot and t' represents the predicted time slot. \\\hline
\textbf{Station} & $i/j$  & any given station in a metro network \\\hline

\textbf{OtD Value (\textbf{OD Value}) }  &  $m^{t}_{i,j} \in \mathbb{R}$  & the number of passengers entering  $i$ at $t$ and destining to $j$  afterward \\\hline

\textbf{OtD Vector (\textbf{OD Vector})} & $D_{i}^{t}=[m^{t}_{i,1}, \cdots, m^{t}_{i,N}]$  & the destination distribution of passengers entering $i$ at $t$  \\\hline

\textbf{Full OD Matrix (OD Matrix)} & $M_{t}=(m^{t}_{i,j})_{N \times N}$ & a matrix composed of all the OD values at $t$ in a metro network \\\hline

\textbf{OD Probability Matrix} & $MP_{t}=(p^{t}_{i,j})_{N \times N}$ & It is the row normalization of full OD Matrix $M_{t}$. $p^{t}_{i,j}=\frac{m^{t}_{i,j}}{\sum_{j=1}^{N}m^{t}_{i,j}}$ and $\sum_{j=1}^{N}p^{t}_{i,j}=1$ \\\hline

\textbf{Finished OD Matrix} & $ MF_{t}=(mf^{t,t'}_{i,j})_{N \times N}$ & a matrix with element $mf^{t,t'}_{i,j} \in \mathbb{R}$ representing the number of passengers entering  $i$ at input time slot $t$ and finishing their trips at $j$  before  output time slot $t'$ ($t$ $<$ $t'$) \\\hline

\textbf{Delayed OD Matrix} & $MD_{t}=(md^{t, t'}_{i,j})_{N \times N}$ & a matrix with element $md^{t,t'}_{i,j} \in \mathbb{R}$ representing the number of passengers entering  $i$ at input time slot $t$ and finishing their trips at  $j$  after output time slot $t'$ ($t$ $<$ $t'$) \\\hline

\textbf{Delayed OD Probability Matrix} & $MDP_{t}=(dp^{t, t'}_{i,j})_{N \times N}$ & It is the row normalization of Delayed OD Matrix $MD_{t}$. $dp^{t, t'}_{i,j}=\frac{md^{t, t'}_{i,j}}{\sum_{j=1}^{N}md^{t, t'}_{i,j}}$ and $\sum_{j=1}^{N}dp^{t, t'}_{i,j}=1$ \\\hline

\textbf{Delayed OD Ratio Matrix} & $MDR_{t}=MD_{t}/M_{t} = (dr^{t}_{i,j})_{N \times N}$ & a matrix with element $dr^{t,t'}_{i,j} \in [0, 1]$ representing the probability of passengers finishing their trips at  $j$  after $t'$ among those entering  $i$ at $t$ ($t$ $<$ $t'$) \\\hline

\textbf{ODt Value} & $\bar m^{t}_{i,j}\in \mathbb{R}$ & the number of passengers entering $i$ before t and arriving at $j$ at $t$ \\\hline

\textbf{ODt Vector} & $\bar D_{i}^{t}=[\bar m^{t}_{i,1}, \cdots, \bar m^{t}_{i,N}]$  & the destination distribution of passengers entering  $i$ before t and arriving at  $j$ at $t$ \\\hline

\textbf{Inflow} & $I_{t} = [\sum_{j=1}^{N} m^{t}_{1,j},\cdots,\sum_{j=1}^{N} m^{t}_{N,j}]$  & a vector with element representing the number of passengers entering all origin stations at $t$  \\\hline

\textbf{Finished Inflow} & $ IF_{t} = [\sum_{j=1}^{N} mf^{t,t'}_{1,j},\cdots,\sum_{j=1}^{N} mf^{t,t'}_{N,j}]$   & a vector with element representing the number of passengers entering all origin stations at $t$ and finishing their journey before $t'$ \\\hline

\textbf{Delayed Inflow}  &  $ID_{t} = [\sum_{j=1}^{N} md^{t,t'}_{1,j},\cdots,\sum_{j=1}^{N} md^{t,t'}_{N,j}]$   & a vector with element representing the number of passengers entering all origin stations at $t$ and finishing their journey after $t'$ \\\hline

\textbf{Outflow} & $O_{t} = [\sum_{i=1}^{N} \bar m^{t}_{i,1},\cdots,\sum_{i=1}^{N} \bar m^{t}_{i,N}]$  & a vector with element representing the number of passengers arriving exit station at $t$  \\

\bottomrule
\end{tabular}
\label{tab:not}
\end{table*}

\section{Preliminary}
\label{lab:Preliminary}
\textbf{Terminology.} Suppose the metro network in our paper has $N$ station, we define some concepts (as shown in Table \ref{tab:not}) to facilitate our discussion.

\textbf{Problem Formulation.} The metro system contains $N$ stations. We aim to predict the OD Matrix $\hat{M}_{t'}$ at the predicted time slot $t'$ with the latest $P$ previous Finished OD matrices $[ MF_{t'-1},\dots,  MF_{t'-P}]$,  other observed trips information $\text{TR}$ (e.g. Inflow, Outflow, ODt Matrix, as defined in Table \ref{tab:not}) and prior knowledge $\text{PK}$ (e.g. geographical distance between stations). The problem can be formulated as follows:
\begin{equation}
\hat{M}_{t'}=f([MF_{t'-1},\dots,  MF_{t'-P}], \text{TR}, \text{PK}, \text{G})
\end{equation}
where $f$ is a deep learning based mapping function and $[t'-{1},\dots, t'-{P}]$ is the input time slot sequence.

\textbf{Loss Function.}
Following previous works \cite{zhang2021short, cheng2021real}, we adopt mean square error (MSE) as the loss function, denoted as: 
\begin{equation}
\begin{aligned}
\mathcal{L}&=MSE (M_{t'}-\hat{M}_{t'}) \\
&= \frac{1}{N \times N} \sum_{i=1}^{N}\sum_{j=1}^{N}(m_{ij}^{t'}-\hat{m}_{ij}^{t'})^2
\end{aligned}
\label{lab:equ1}
\end{equation}
To large OD values, the difference between their ground truth and prediction is enlarged by MSE. Thus, the model is trained toward reducing the predicting error of large OD values instead of small OD values. This is consistent with the real-world demand that large OD volumes are more important and should be paid more attention in metro scenario.

\section{Methodology}
In this paper, we propose a Completion based Adaptive Heterogeneous Graph Convolution Spatiotemporal Predictor (C-AHGCSP) to make full use of the real-time and historical observed passenger flows (e.g. finished OD matrix sequence, inflow/outflow) as well as the factors affecting the global mobility pattern correlations between stations to predict the OD Matrix in the target time slot.

\begin{figure}[htb]
\centering
\includegraphics[width=0.480\textwidth, height=0.25\textheight]{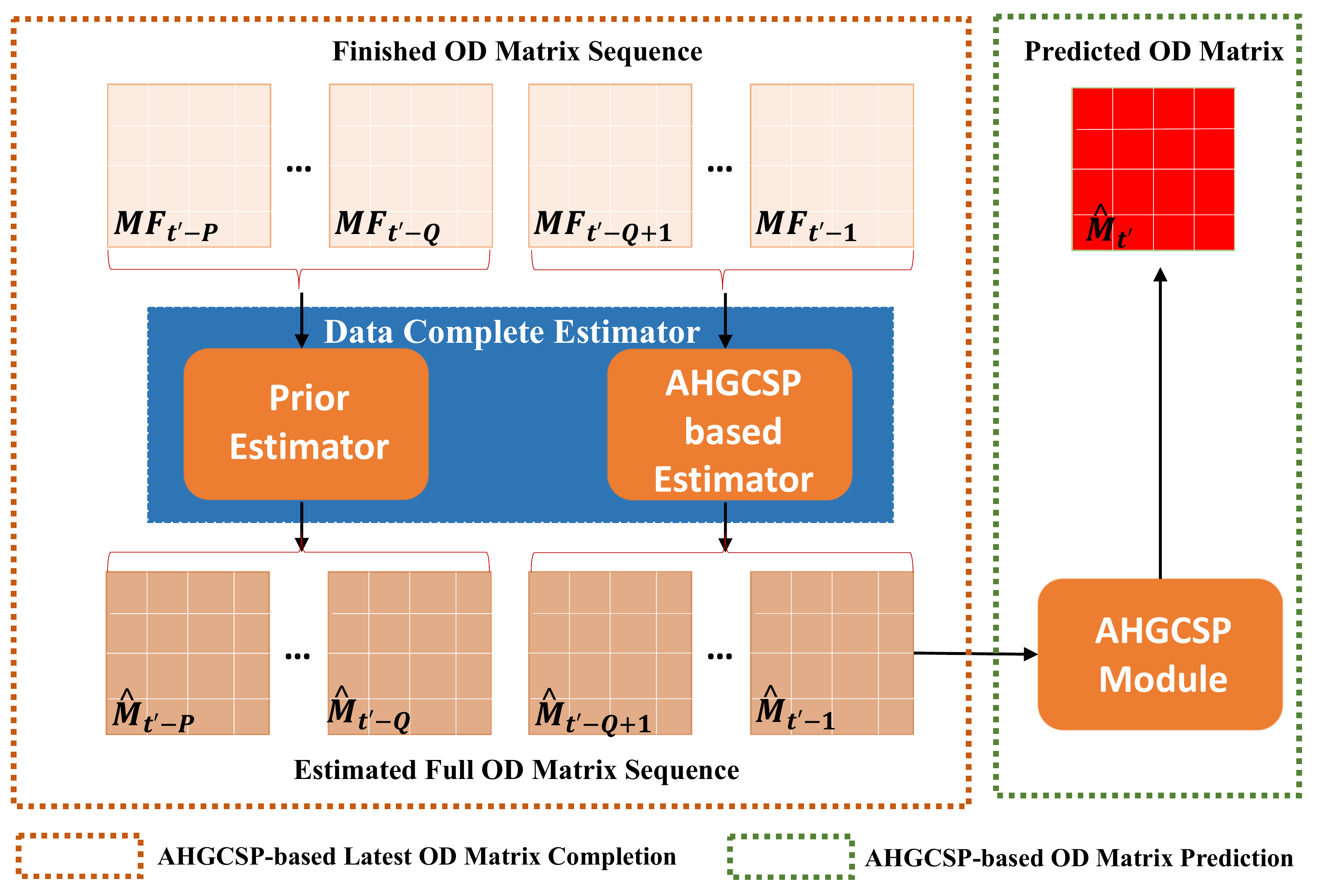}
\caption{Our model is called C-AHGCSP and is composed of AHGCSP-based Latest OD Matrix Completion and AHGCSP-based OD Matrix Prediction. The former is utilized to estimate the latest OD matrix sequence by two estimators (i.e. Prior Estimator and AHGCSP based Estimator) and the latter is leveraged for OD matrix prediction at the next time step. Both of them have the module AHGCSP (as shown in Figure \ref{fig:ahg}) and it is in charge of modeling the various passenger spatiotemporal mobility to predict their destination distribution. }
\label{fig:mod}
\end{figure}

The framework of C-AHGCSP  is shown in Figure \ref{fig:mod}, which contains two modules: AHGCSP-based OD matrix completion module and AHGCSP-based OD matrix prediction module. The two modules share same sub-module AHGCSP. Given a target time slot, AHGCSP takes the OD passenger matrixes before as input, learning the spatiotemporal mobility correlations between stations, and output the OD matrix in the target time slot. Be note the target time slot may be the predicted time slot $t'$ or a latest time slot where the OD matrix is incomplete.

The completion module aims to complete latest OD matrixes before the predicted target time slot $t'$. Given a latest time slot with incomplete OD matrixes, it uses AHGCSP and prior knowledge about time cost between two stations to estimate the destination distribution of unfinished trips.

The prediction module takes the time series of observed full OD matrixes and completed OD matrixes before the target time slot $t'$ as input, and use AHGCSP to predict the final OD matrix in the target time slot.


\begin{figure}[htb]
	\centering
	\includegraphics[width=0.49\textwidth, height=0.35\textheight]{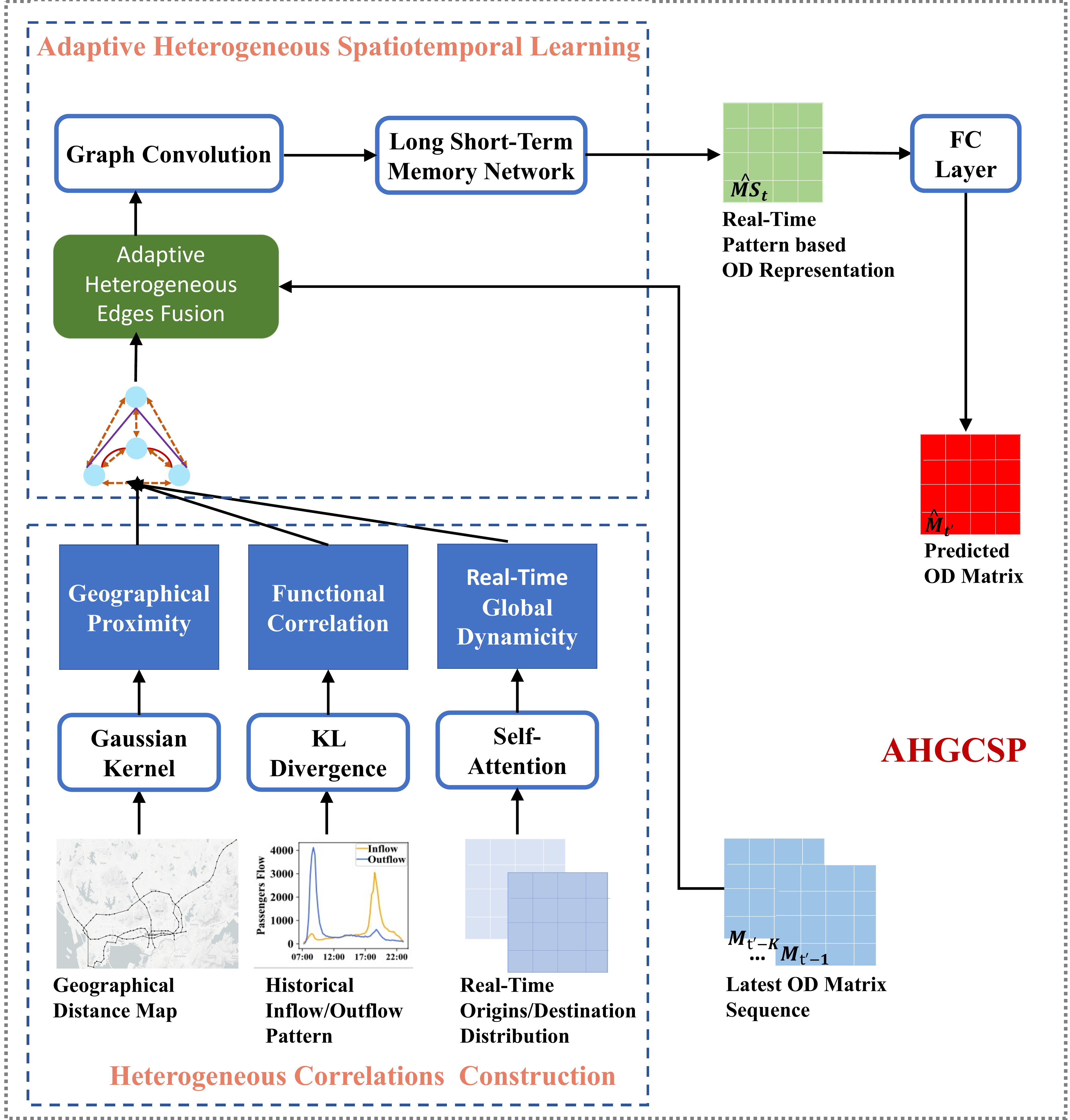}
	\caption{The  AHGCSP module aims to capture the passenger travel motivation from spatiotemporal perspective. It mainly contains two modules. Heterogeneous Correlations Construction module models three passenger motivation correlations among stations from spatial perspective.  Adaptive Heterogeneous Spatiotemporal Learning module takes the latest OD matrix sequence as input and digs out the real-time passenger travel spatiotemporal pattern mainly by GCN and LSTM, outputting a real-time pattern based OD matrix.}
	\label{fig:ahg}
\end{figure}

\subsection{AHGCSP-based OD Matrix Prediction}
This section aims to predict OD matrix at time slot $t'$. More specifically, it predicts the destination distributions of passengers starting at each station (geographical location, region function, etc). For that, we first construct a dynamic station graph with heterogeneous edges based on the OD matrix at each latest time slot.
Then, we adopt an adaptive spatiotemporal learning method to learn the global mobility spatiotemporal correlations from the graph sequence, and output the OD matrix at time slot $t'$.

\begin{figure}[h]
	\centering
	\includegraphics[width=0.43\textwidth, height=0.2\textheight]{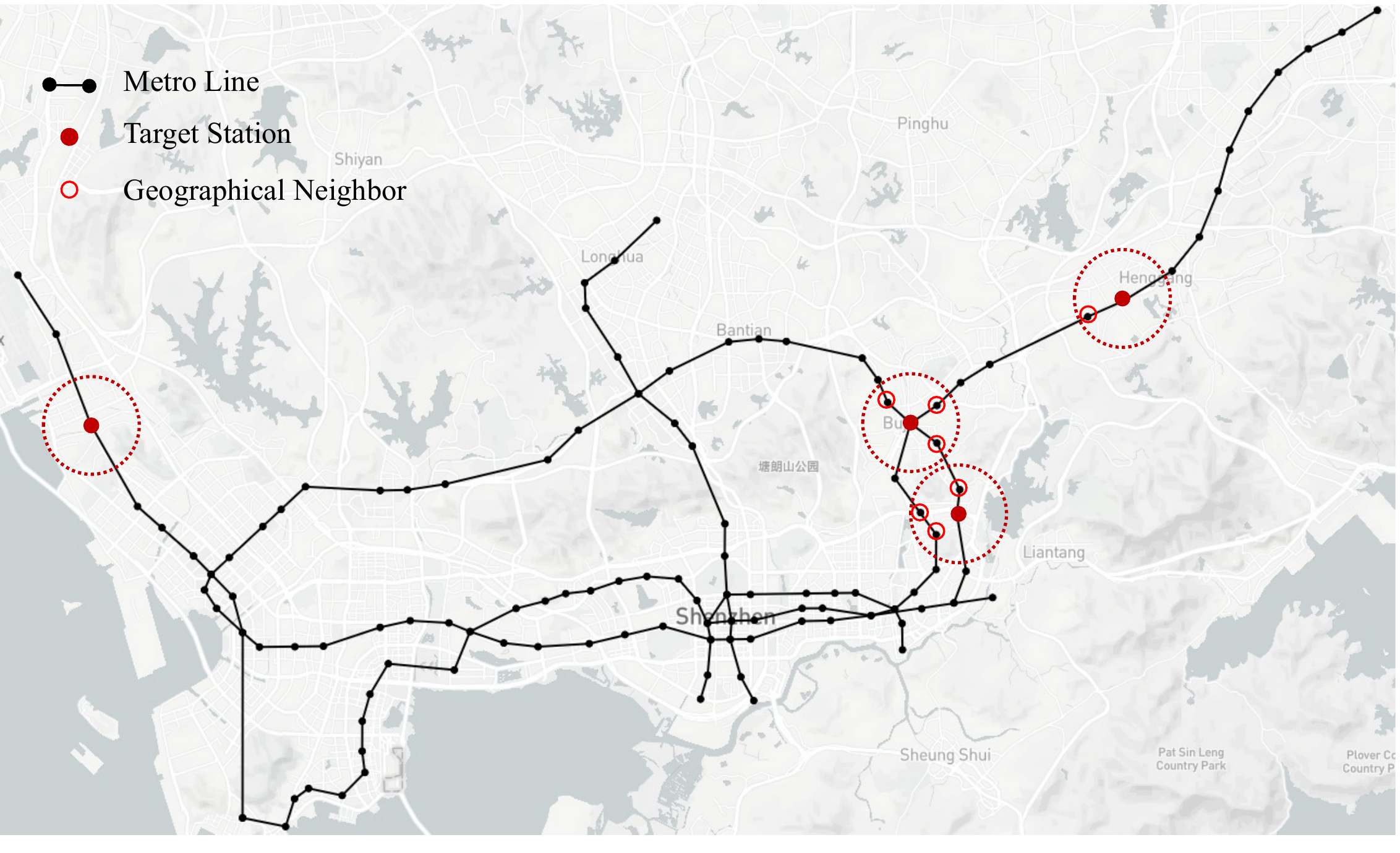}
	\caption{We measure the geographical proximity between any two stations in the metro network based on geographical distance instead of connectivity. Take Shenzhen subway for example, we choose the same empirical distance radius $\text{D}$  for all stations. The stations inside the circle of the target station are its geographical neighbors. The sparsity of the geographical neighbors depends on the empirical threshold $\text{D}$. Some target stations even don't have geographical neighbors.}
	\label{fig:geo}
\end{figure}

\begin{figure*}[htb]
	\centering
	\includegraphics[width=0.98\textwidth, height=0.2\textheight]{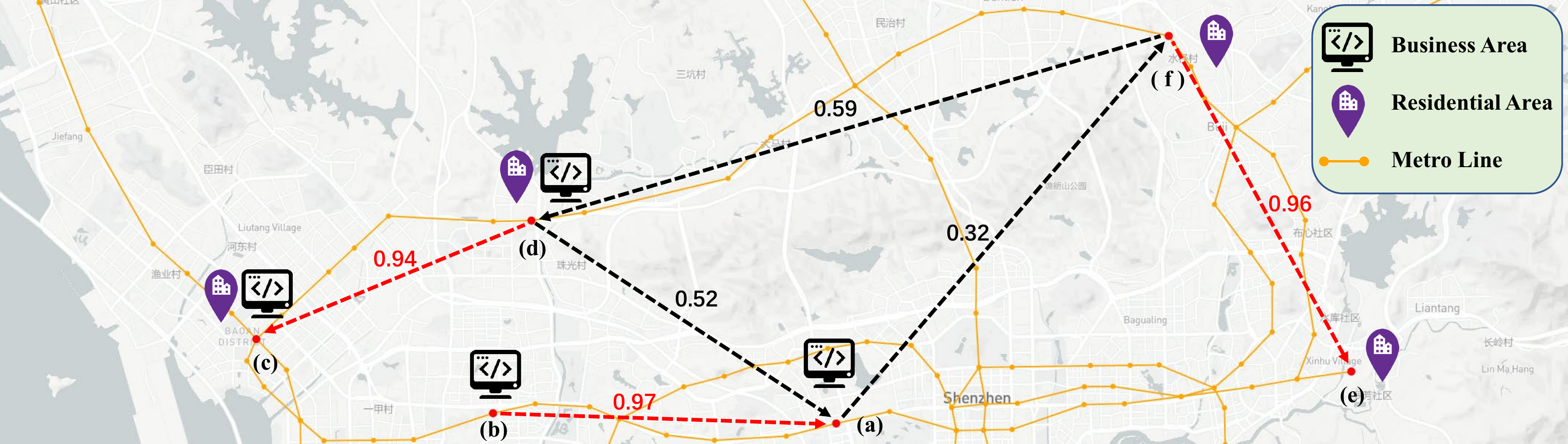}
	\caption{We randomly choose some metro stations to calculate their function similarity score by the KL divergence. As we can see, the metro stations with similar functions have a high function similarity score while the stations with different functions have a low similarity score which proves the effectiveness of KL divergence in reflecting the function similarity. For example, both stations (a) and (b) are located in business areas and their similar score is 0.97 while the score between (a) and station (f) located in residential area is only 0.32.}
	\label{fig:sim}
\end{figure*}

\begin{figure}[h]
	\centering
	\subfigure[Chegongmiao]{
		\includegraphics[width=0.22\textwidth,height=0.13\textheight]{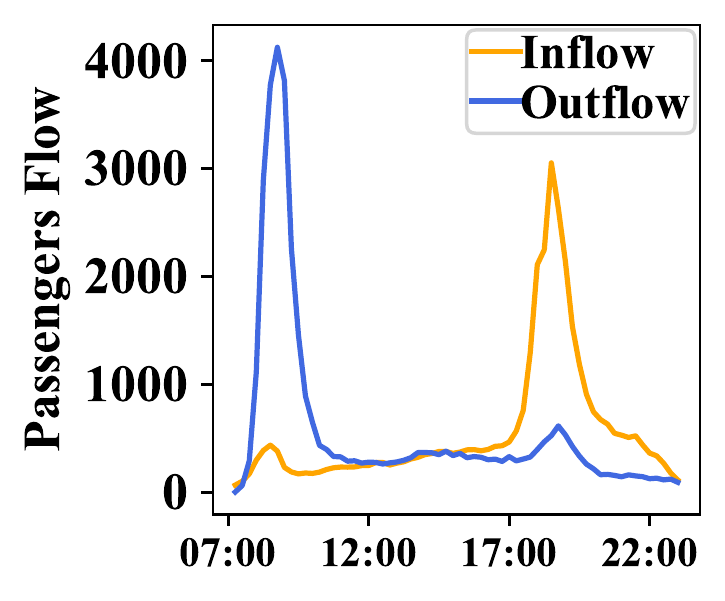}
	}
	\subfigure[Shenda]{
		\includegraphics[width=0.22\textwidth,height=0.13\textheight]{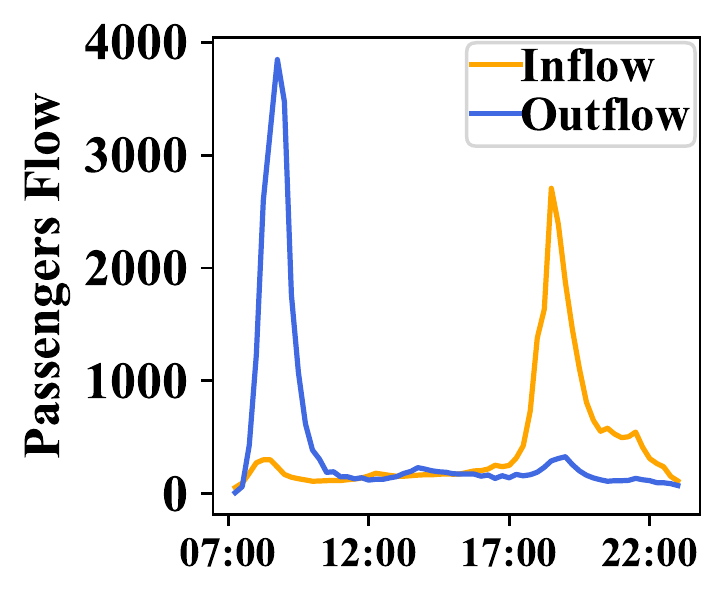}
	}
	\subfigure[Xinxiu]{
		\includegraphics[width=0.22\textwidth,height=0.13\textheight]{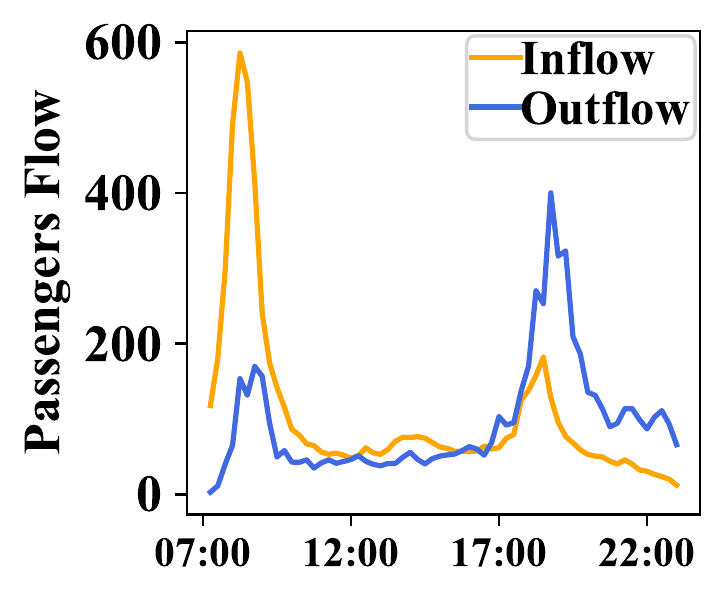}
	}
	\subfigure[Shangshuijing]{
		\includegraphics[width=0.22\textwidth,height=0.13\textheight]{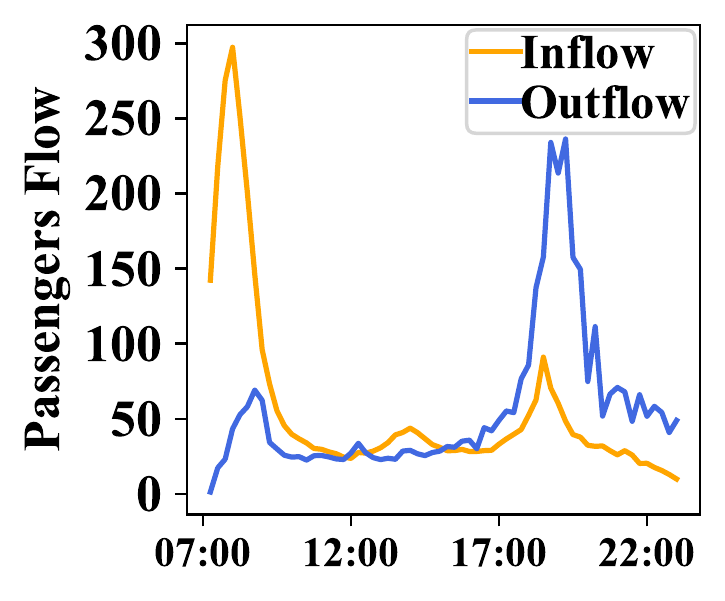}
	}
	\caption{These Figures above show the inflow and outflow patterns on historical Monday of Shenzhen subway stations with different region functions: Chegongmiao and Shenda both locate in business areas and they share similar single peak in inflow (outflow) pattern. Xinxiu and Shangshuijing both in residential areas also share similar flow patterns. However, Chegongmiao and Xinxiu show almost reverse inflow and outflow patterns because they have different region functions.}
	\label{fig:pat}
\end{figure}

\begin{figure}[htb]
\centering
\includegraphics[width=0.45\textwidth, height=0.14\textheight]{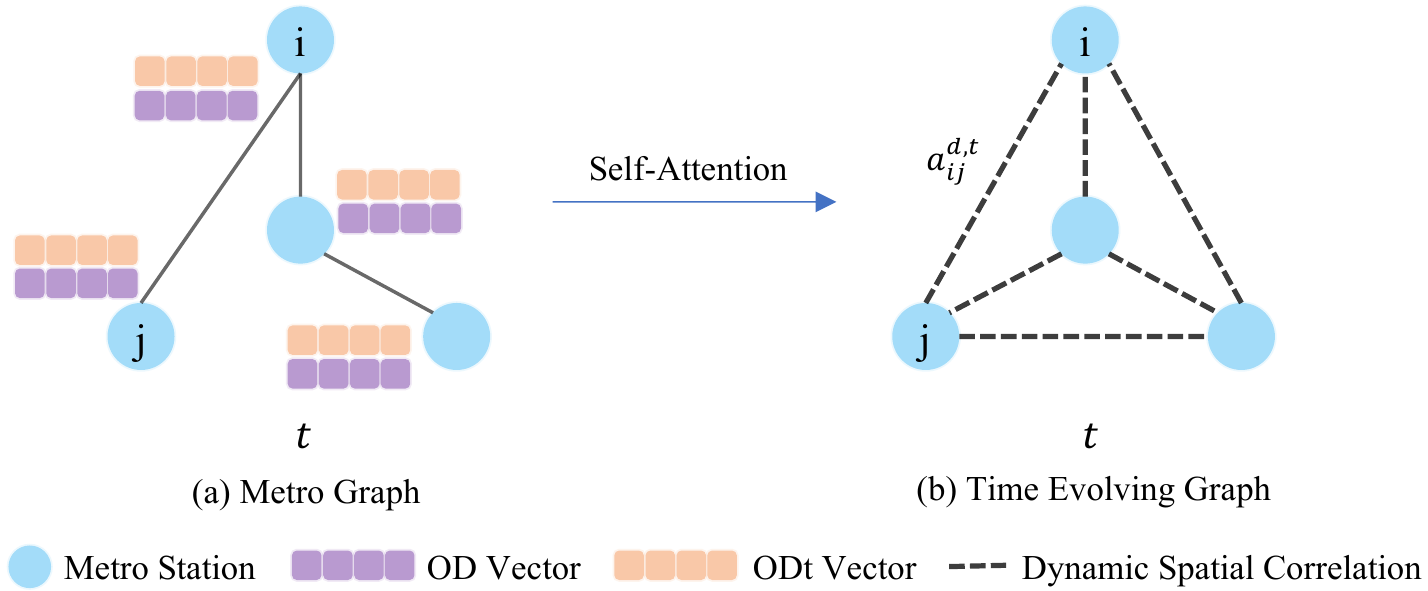}
\caption{All the stations are connected to each other dynamically and asymmetrically based on the similarity of the current destination distributions. $\alpha_{ij}^{dt}$ is the dynamic score to measure the dynamic similarity between station $i$ and station $j$ at time slot $t$.}
\label{fig:dyn}
\end{figure}

\subsubsection{\textbf{Heterogeneous Spatial Correlations Construction}}
We analyze the data and find that some stations share similar passenger travel pattern due to their similar attributes, such as geographical proximity, functional similarity and global dynamicity(as shown in Figure \ref{fig:pat}). That is beneficial for their passenger destination distribution prediction. We model such mobility correlation from three perspectives (i.e. geographical proximity, functional similarity and global dynamicity) by appropriate methods based on the available real-time and historical flow information as well as prior geographical information.

\textbf {Gaussian Kernel based Geographical Proximity}: Passengers in geographical close areas tend to share more similar destinations than those in distant areas \cite{ye2020build, wang2019origin}. In this perspective, the closer the stations are, the more they are likely to learn from each other in passenger travel pattern. Inspired by \cite{wang2019origin}, we define the geographical neighbor of station $i$ (as shown in Figure \ref{fig:geo}) as follows:
\begin{equation}
\mathcal{N}_{i}=\left\{s_{j} \mid \operatorname{geo}\left(s_{i}, s_{j}\right) \leq \text{D} \right\}
\end{equation}
where $\operatorname{geo}\left(s_{i}, s_{j}\right)$ is the geographical distance calculated by longitude and latitude of each metro station. $\text{D}$ is a distance threshold to determine the range of neighborhood. Further, we measure the geographical proximity between stations using the threshold Gaussian Kernel \cite{LiYS018} as follows:
	
\begin{equation}
\alpha^{g}_{i j}=\left\{\begin{array}{ll}\exp \left(-\frac{\operatorname{geo}(s_{i}, s_{j})^2}{\sigma^{2}}\right), & i = j \text { or } i \in \mathcal{N}_{i} \\ 0, & i \neq j \text { and } i \notin \mathcal{N}_{i}\end{array}\right.
\end{equation}
where $\sigma$ is the standard deviation of the distance between all stations and their neighbors and $\alpha^{g}_{i j} \in [0, 1]$.

\textbf {KL Divergence based Functional Correlation}:
The region that a metro station located in usually has some functions, e.g. business, education, residential, entertainment function and these functions have a large impact on the mobility pattern of passengers in this region \cite{zhao2021mdlf,  liu2020physical}. Metro stations in regions with similar functions are likely to have similar passenger travel patterns, otherwise different patterns. Such similarity is beneficial for prediction. According to our observation from the collected data, the historical inflow and outflow patterns of stations within a day  can largely distinguish the region functions (as shown in Figure \ref{fig:pat}). Therefore, we novelly propose to measure the function similarity between any two stations based on flow patterns by KL divergence.

Specifically, we sum up the inflow of all the historical days with the same week attribute (e.g. Monday) to obtain a $T$ dimension inflow (outflow) vector for each station $i$, denoted as $IV_{i}=[iv_{i}^{1},iv_{i}^{2},\cdots,iv_{i}^{T}]$, where $iv_{i}^{t}$ represents the total number of passengers entering station $i$ at $t$ historically. To eliminate the influence of passenger volume scale on region function, the inflow vector is normalized into a probability distribution as $p_{i}=[\frac{iv_{i}^{1}}{\sum_{t=1}^{T}iv_{i}^{t}},\frac{iv_{i}^{2}}{\sum_{t=1}^{T}iv_{i}^{t}},\cdots,\frac{iv_{i}^{T}}{\sum_{t=1}^{T}iv_{i}^{t}}]$. We measure the similarity of inflow probability distributions between any two stations by KL divergence as follows:
\begin{equation}
SI_{ij}=1-D(p_{i} \| p_{j})=1-\sum_{t=1}^{T} p_{i}^{t}\log \frac{p_{i}^{t}}{p_{j}^{t}}
\end{equation}
where $D(p_{i} \| p_{j})$ is the KL divergence score and $SI_{ij}$ is the similarity score of inflow pattern between metro station $i$ and metro station $j$. The more similar the inflow patterns of two stations is, the larger the $SI_{ij}$ is. Similarly, the similarity score of outflow pattern between metro station $i$  and $j$ is calculated and denoted as $SO_{ij}$. We fuse $SI_{ij}$ and $SO_{ij}$ to measure the function similarity between two stations as follows:
\begin{equation}
\alpha_{i j}^{f}= wSI_{ij}+(1-w)SO_{ij}
\end{equation}
where $\alpha_{i j}^{f}$ is the function similarity score between station $i$ and station $j$. The larger $\alpha_{i j}^{f}$ is, the more similar functions the two stations share. $w \in \mathbb{R}$ is the coefficient to measure the contribution of inflow pattern to the function similarity. This coefficient is trainable. We randomly choose some stations to validate the effectiveness of KL divergence in Figure \ref{fig:sim}.

\textbf {Self-Attention based Real-Time Global Dynamicity}:
The geographical graph and functional graph above are primarily based on prior knowledge and historical data. They reflect static spatial correlations in the metro network. The dynamic spatial correlations based on real-time passenger flow are also crucial for prediction.  They contain real-time mobility information which is not contained in predefined graphs.  For example, a sudden abnormal event like vehicle defects of some stations is likely to change the passenger mobility of many related stations currently and such impact might last for some time. Therefore, we propose to capture the time evolving spatial correlations in the whole network based on real-time destination distribution information, i.e. OD Vector and ODt Vector (defined in Table \ref{tab:not}). Inspired by the self-attention mechanism in transformer \cite{vaswani2017attention}, we design the dynamic attention score between any two stations as follows:

\begin{equation}
\begin{aligned}
e_{{i}{j}}^{t}& = f_{key}(D_{i}^{t})^{T} \boldsymbol{\cdot} f_{query}(D_{j}^{t})\\
\bar e_{{i}{j}}^{t} & =  \bar f_{key}(\bar D_{i}^{t})^{T} \boldsymbol{\cdot} \bar f_{query}(\bar D_{j}^{t})\\
\alpha_{ij}^{dt} &= w \frac{\exp (e_{ij}^{t})}{\sum_{j=1}^{N} \exp (e_{ij}^{t})} + \bar w \frac{\exp (\bar e_{ij}^{t})}{\sum_{j=1}^{N} \exp (\bar e_{ij}^{t})}
\end{aligned}
\end{equation}
where $f_{key}$, $f_{query}$, $\bar f_{key}$, $\bar f_{query}$ are all three fully connected layers with activation function ReLu. $e_{{i}{j}}^{t}$ represents the real-time passenger destination distribution similarity between two stations where passengers entering stations at $t$. $\bar e_{{i}{j}}^{t}$ represents the real-time passenger destination distribution similarity between two stations where passengers entering stations before $t$ and leave the metro network at $t$. $w$ and $\bar w$ are the trainable coefficients to attach different weights to these two similarity. $\alpha_{ij}^{dt}$ is the final dynamic score (as shown in Figure \ref{fig:dyn}).



\subsubsection{\textbf{Adaptive Heterogeneous Spatiotemporal Learning}}
Up to now, three heterogeneous relations have been constructed for any station pair $ij$ at time slot $t$, i.e. $\alpha_{i j}^{dt}, \alpha_{i j}^{f}, \alpha_{i j}^{g}$ to reflect the complex passenger travel motivation correlations among stations sufficiently. Intuitively, the correlation strength of any station pair depends on all three relations but each relation contributes differently and we assign different weights to them to obtain the fused correlation score as follow:
\begin{equation}
\begin{aligned}
\alpha_{i j}^{t} = w_{d} \alpha_{i j}^{dt} + w_{f}\alpha_{i j}^{f} + w_{g}\alpha_{i j}^{g}\\
h_{i j}^{t} =\frac{\exp (\alpha_{ij}^{t})}{\sum_{j=1}^{N} \exp (\alpha_{ij}^{t})}
\label{equ:ada}
\end{aligned}
\end{equation}
where $h_{i j}^{t} \in [0, 1]$ is the normalization of $\alpha_{ij}^{t}$.

\textbf{Graph Convolution based Heterogeneous Correlation Extraction}
After the correlations among stations have been constructed and fused adaptively based on the spatiotemporal attributes, each station needs to learn useful destination distribution information from related stations. Recently, Graph Convolution Network (GCN) has achieved the state-of-the-art performance in many traffic tasks, such as metro passenger prediction \cite{9207049}, taxi demand prediction \cite{geng2019spatiotemporal}. Following previous works \cite{wang2019origin}, we utilize graph convolution to aggregate the useful passenger mobility information from all other stations for the target station $i$ based on their fused correlation to produce its high-level mobility pattern representation at each input time slot $t$ as $Y^{l}_{i,t} = \boldsymbol{\rho}((\sum_{j=1}^{N} h_{i j}^{t}D_{j}^{t})W^{l}+b^{l})$  $Y^{l}_{i,t}$ is the $l_{th}$ layer feature of station $i$ at time slot $t$. $D_{j}^{t}$ is the destination distribution of station $j$ at time slot $t$. $\boldsymbol{\rho}$ is the activation function (i.e. ReLU in this paper) and $W^{l}, b_{l}$ are the learnable parameter at layer $l_{th}$. Following previous works \cite{kipf2016semi}, to alleviate the over-smoothing problem, there are only two graph convolutional layers in our model.

All the stations and time slots share the same graph convolution parameters, we rewrite the related operations as matrix form as $G_{t}=\text{GCN}(H_{t}, M_{t})$ where $H_{t}=(h_{i j}^{t})_{N \times N}$ is the fusion matrix for the whole network at time slot $t$ and $M_{t}=[D_{1}^{t}, \cdots, D_{N}^{t}]$ is the OD Matrix at time $t$, namely the passenger destination distributions of all the stations. $G_{t}\in \mathbb{R}^{N \times G}$ is the output of the GCN.

\textbf{LSTM based Temporal Dependency Extraction}
We have extracted the heterogeneous correlations on the metro network at each time slot $t$,  represented by $G_{t}$. Then we need to dig out the consecutive time dependency from latest high-level destination distribution sequence $[G_{t'-1},\dots, G_{t'-P}]$. Following previous works \cite{wang2019origin}, Long Short Term Memory Network (LSTM) is utilized to capture the real-time passenger mobility pattern as $\hat{MS}_{t'} = \text{LSTM}([G_{t'-1},\dots, G_{t'-P}])$ where $\hat{MS}_{t'} \in \mathbb{R}^{N \times U}$ is the output containing short-term passenger travel pattern information. Afterward, $\hat{MS}_{t'}$ is fed into one fully connected layer with ReLU activation to obtain the final prediction as follows:
\begin{equation}
\hat{M}_{t'} = ReLU(\hat{MS}_{t'}W_{1}+b_{1})
\end{equation}
where $[W_{1},b_{1}]$ are trainable parameters and $W_{1} \in \mathbb{R}^{U \times N}$.


\begin{figure*}[h]
	\centering
	\includegraphics[width=1.00\textwidth, height=0.2\textheight]{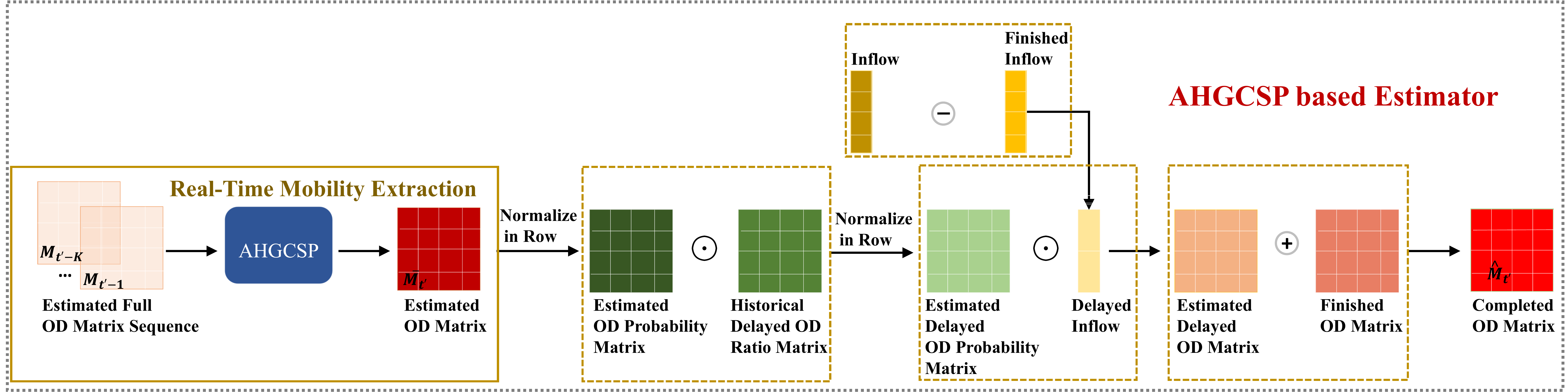}
	\caption{The AHGCSP based Estimator is responsible for full OD matrix completion at each target input time slot $t'$ by taking advantages of all the available data sufficiently, i.e. an estimated full OD matrix sequence at previous input time slots, inflow, finished inflow, finished OD matrix at the target input time slot as well as the historical delayed OD probability matrix. The definitions of all the terminologies in the figure above is listed in Table \ref{tab:not}.}
	\label{fig:est}
\end{figure*}


\begin{figure}[h]
	\centering
	\subfigure[Incompleteness Analysis on Shanghai Dataset]{
		\includegraphics[width=0.38\textwidth,height=0.15\textheight]{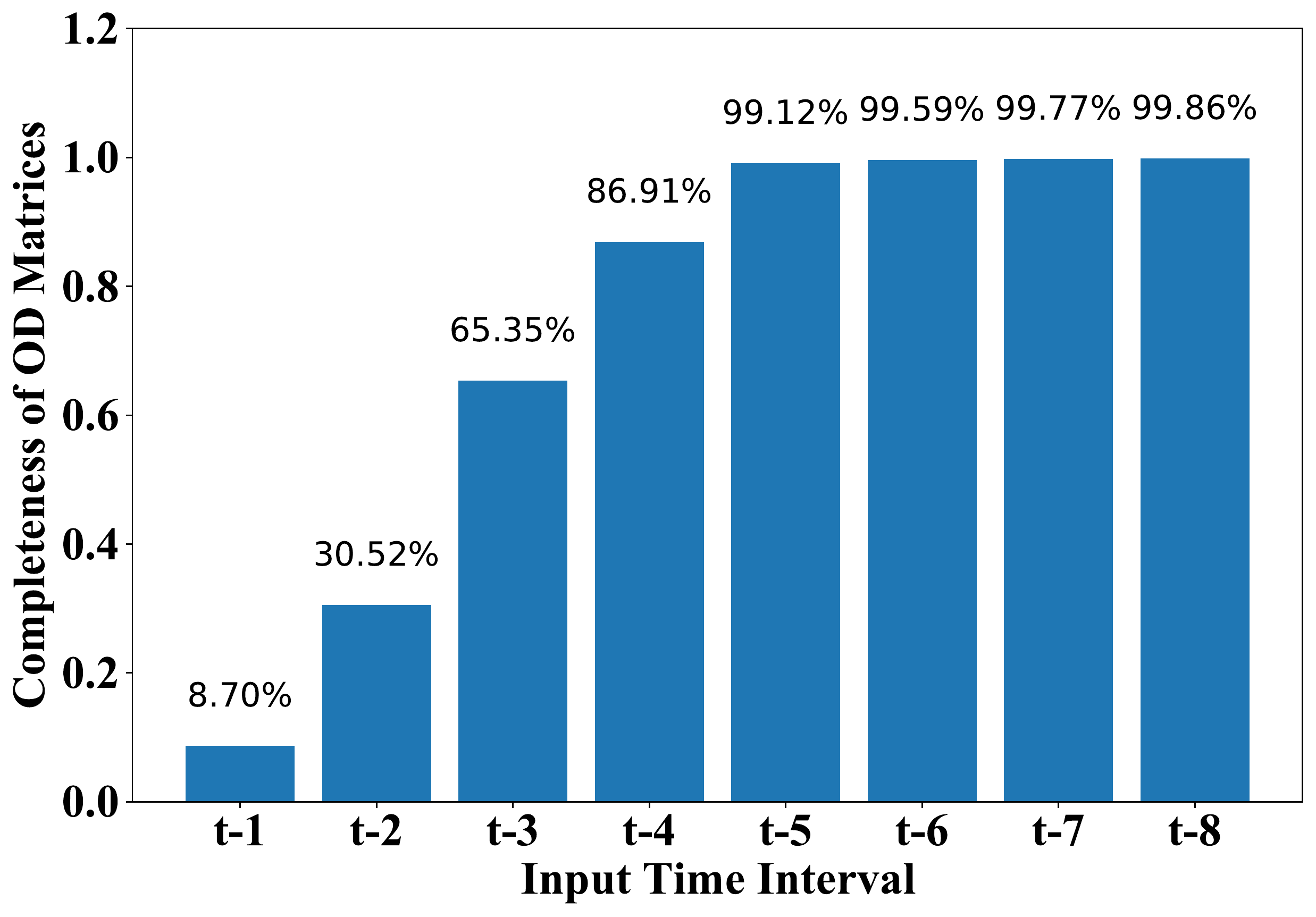}
	}
	\subfigure[Incompleteness Analysis on Shengzhen Dataset]{
		\includegraphics[width=0.38\textwidth,height=0.15\textheight]{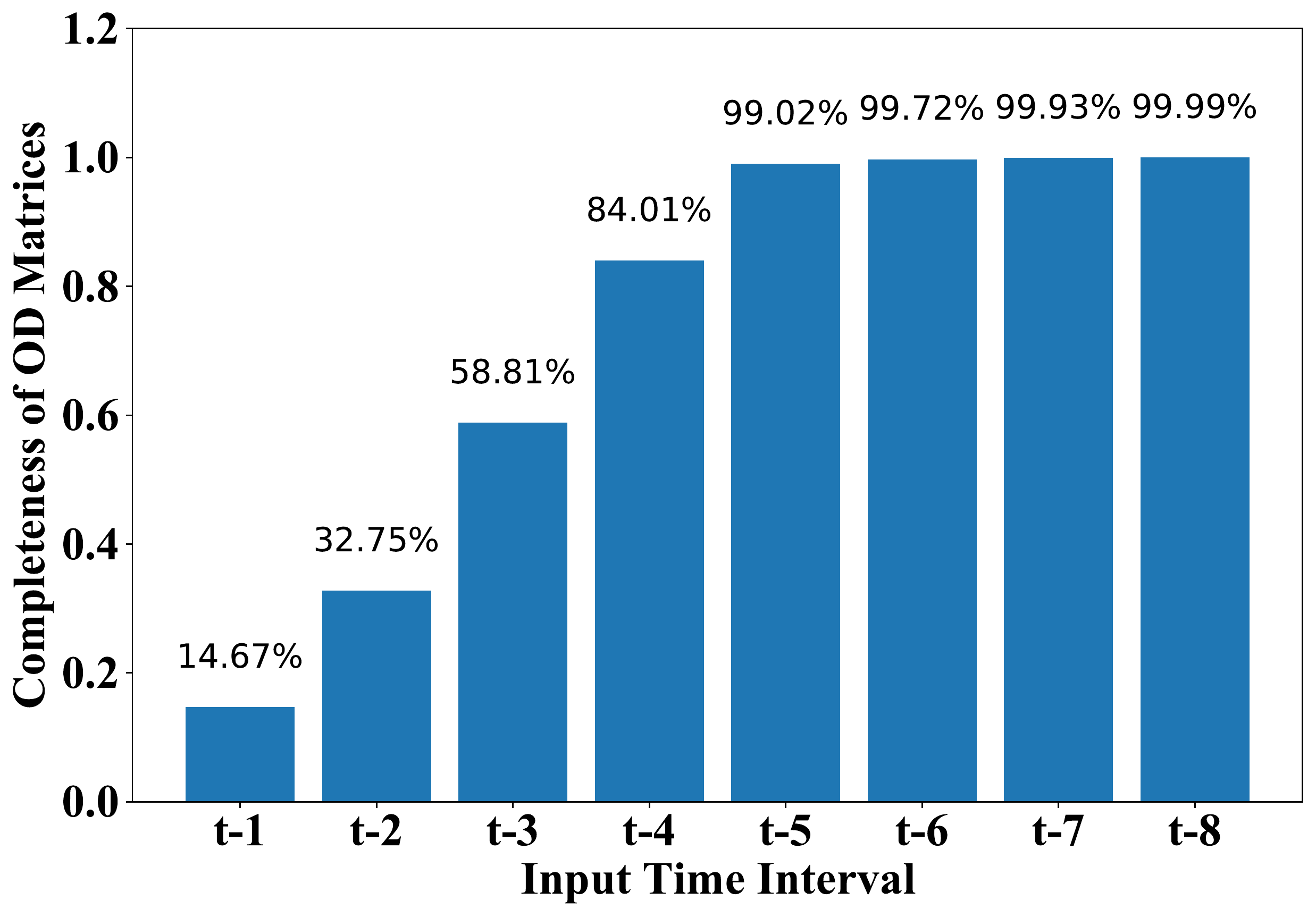}
	}
	\caption{We generate the data samples with eight input time slots and analyze the completeness of each  input time slot. $t'$ is the target time slot. The time granularity is 15 minutes. It can be observed that the closer one matrix is near the predicted matrix, the more incomplete it is. Note that the matrices after $[t'-4]$ is almost complete due to the fact that most trips are finished within one hour in metro network and few trips require travel time more than 75 minutes.}
	\label{fig:incom}
\end{figure}

\subsection{AHGCSP-based Latest OD Matrix Completion}
\label{sub:full}
Unlike the ride-hailing OD prediction task, in the metro scenario, the finished OD matrix at each input time slot is likely to be incomplete due to the unfinished trips before the predicted time slot and the full OD matrix is unavailable. The incompleteness leads to the sparsity of finished OD matrix.
We expect the latest mobility information to improve the model performance significantly in metro network just like other scenario, especially for real-time abnormal situations caused by special events.
Different from the previous studies which fed the incomplete OD matrix sequence or directly into the model \cite{gong2020online, cheng2021real, zhang2021short}, we novelly propose a data complete estimator to estimate a full OD matrix.  Specifically, the data complete estimator is composed of a prior estimator and an AHGCSP based estimator. It turns the input data, i.e. the finished OD matrix sequence $[MF_{t'-1},\dots,  MF_{t'-P}]$ at latest $P$ time slots into an estimated full OD matrix sequence $[\hat M_{t'-1},\dots,  \hat M_{t'-P}]$.

\begin{figure}[htb]
\centering
\includegraphics[width=0.4\textwidth, height=0.2\textheight]{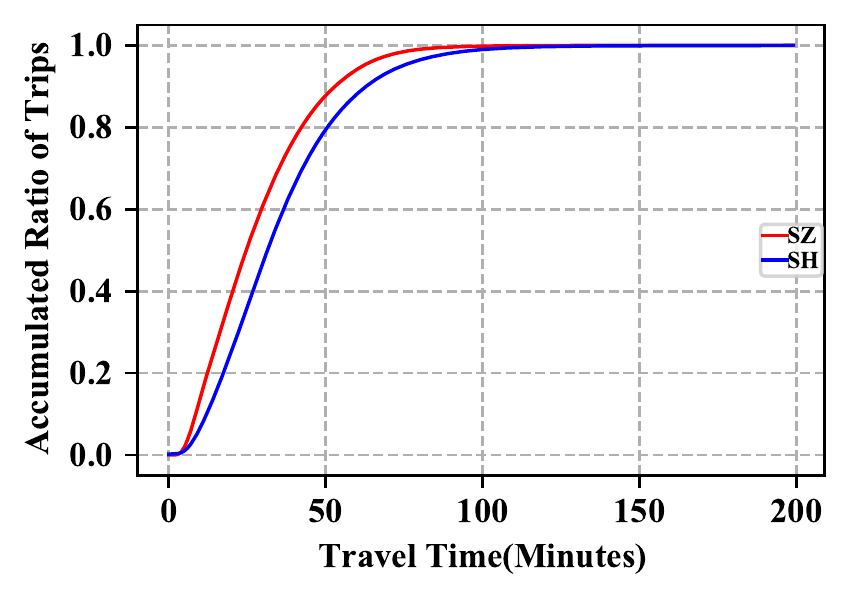}
\caption{This figure shows accumulated travel time distribution of all the historical trips.}
\label{fig:tra}
\end{figure}

\subsubsection{\textbf{Prior Estimator}}
According to the observations of our data (as shown in Figure \ref{fig:incom}), we find that not all the finished OD matrices at the input time slots are incomplete, i.e. some finished OD matrices are complete or nearly complete. The incompleteness of a finished OD matrix is determined by the time gap (i.e. time distance between input time slot and output time slot) and the passengers travel time. We propose to measure the completeness of an finished OD matrix as follows:
\begin{equation}
C_{t}=\frac{\sum MF_{t}}{\sum M_{t}} = f((t'- t-1)\delta, (t' - t)\delta, TR_{t})
\end{equation}
where $C_{t}\in [0, 1]$ measures the matrix completeness. $MF_{t}$ is the finished OD matrix and $M_{t}$ is the full OD matrix. $t$ is the input time slot and $t'$ is the output time slot. $g$ is the time difference between $t'$ and $t$ and $g \in [1, \cdots, P]$. $\delta$ is time granularity. $(t'- t-1)\delta/(t' - t)\delta$ is the minimum/maximum travel time that a passenger entering the metro network at $t$ has before target time slot $t'$. $TR_{t}=\{tm_{t}^{k}|k=1,\cdots, K\}$ is a set of travel time of all trips starting at $t$ with $tm_{t'}^{k}$ denoted as travel time of one trip.

For each passenger trip, the travel time is influenced by many factors, such as its chosen route of the OD pair, the waiting time and the time on train vehicles. However, we can expect it to have a maximum value and a minimum value because there exists shortest travel path between OD pair and the time that passengers willing to spend on trains is limited, i.e. $TR_{t}\in [\text{MIN}_{TR_{t}}, \text{MAX}_{TR_{t}}]$. And if $\text{MIN}_{TR_{t}} \geq (t' - t)\delta$, $C_{t}=0$ because no passenger has finished their trips. Reversely, if $\text{MAX}_{TR_{t}} \leq (t' - t-1)\delta$, $C_{t}=1$ because all passengers have finished their journeys before $t$. If both the conditions above are not satisfied, $C_{t} \in (0, 1)$. Therefore, to obtain complete matrix, the input time slot should satisfy $t \leq t'-(1+\frac{\text{MAX}_{TR_{t}}}{\delta}) \leq t'-Q$ where $Q= \lfloor \frac{\text{MAX}_{TR}}{\delta}+1 \rfloor$. If $P > Q$, finished matrices $[MF_{Q}, \cdots, MF_{P}]$ are complete matrices. Therefore, we can estimated the full OD matrices from $[t'-Q, \cdots, t'-P]$ as follows:
\begin{equation}
\hat M_{t} = MF_{t}, t \in [t'-Q, \cdots, t'-P]
 \end{equation}

Empirically, as shown in Figure \ref{fig:tra}, almost all the trips are finished within 100 minutes. According to the statistics, there are 94.2\% of trips on Shenzhen dataset and 88.2\% of trips on Shanghai dataset finished within 60 minutes. Another observation of incompleteness of finished OD matrices is shown in Figure \ref{fig:incom}, finished OD matrices before $t'-5$ are incomplete but those ranging from $[t'-5, \cdots, t'-8]$ are nearly complete at both datasets. Therefore, we set $Q=5$ based on our datasets.

\subsubsection{\textbf{AHGCSP based Estimator}}
In this section, we estimate the full OD matrix of each time slot $t \in [t'-1, \cdots, t'-Q+1]$ before the predicted time slot $t'$. The full OD matrix can be partitioned and formulated as follows:
\begin{equation}
\footnotesize
\left\{\begin{array}{ll}
M_{t} &= MF_{t} + MD_{t} \\
MD_{t} &= MDP_{t} * ID_{t}\\
ID_{t} &= I_{t} - IF_{t}
\end{array}\right.
\Rightarrow M_{t} = MF_{t} +  MDP_{t} * (I_{t} - IF_{t})
\label{equ:m}
\end{equation}
where $MD_{t}$ is the Delayed OD Matrix, $MDP_{t}$ is the Delayed OD Probability Matrix. $ID_{t}$ is the Delayed Inflow. All the terminologies have been defined in Table \ref{tab:not}. Since the finished OD matrix $MF_{t}$, Inflow $I_{t}$, Finished Inflow $IF_{t}$ are all available, to estimate $M_{t}$, we just need to estimate $MDP_{t}$.

A simplest way is to use the $MDP_{t}$ at the same time slot at previous week for estimation, which is denoted as $MDP_{t}^{W}$. However, such estimation only contains the historical mobility information, no real-time mobility information. On the other hand, the OD matrix sequence at time slots $[t'-Q, \cdots, t'-P]$ is almost complete and already available which contains real-time mobility information and our module AHGCSP can learn from the sequence for $MDP_{t}$ estimation.

There may be multiple latest time slots where the OD matrixes are incomplete. The closer of a latest time  slot $t$ to the predicted target time slot $t'$, the higher incompleteness degree of the OD matrixes. Our strategy is to use an iterative completion method as follow: we first complete the OD matrix at the first time slot with the lowest loss rate through combining AHGCSP and prior prior knowledge of time cost between two stations to estimate the destination distribution of unfinished trips. Afterwards, it uses past completed OD matrixes as input to complete the OD matrix in the second slot, and so on until all latest OD matrixes are completed.

More specifically, AHGCSP is used to take an estimated full OD matrix sequence at previous $K$ time slots as input, defined as $\hat S_{t}=[\hat M_{t-1}, \cdots, \hat M_{t-K}]$ and output a predicted OD matrix at time slot $t$, defined as $\hat M_{t}$. If we normalize $\hat M_{t}$ for each row, we can obtain the destination distribution of all trips starting from each station, called OD probability matrix $MP_{t}$ as defined in Table \ref{tab:not}. If we  further take the time cost into consideration, we can estimate the destination ratio of unfinished trips. Given the number of passengers of an O-D pair at time slot $t$, the time cost determine the ratio of passengers finished or unfinished before time slot $t'$, denoted as  Delayed OD Ratio Matrix $MDR_{t}$ (defined in \ref{tab:not}). Since the travel time is relatively stable in a metro system, we utilize historical trips to calculate $MDR_{t}$. 

Based on the above analysis, the $MDP_{t}$ is calculated as follows:
\begin{equation}
\footnotesize
\hat MDP_{t} = f_{_{N}}(f_{_{N}}(AHGCSP( \hat S_{t}) * MDR_{t}^{W})
\label{equ:mdp}
\end{equation}
where $f_{_{N}}$ is the normalization operation in row.

After we obtain estimation of $MDP_{t}$, we rewrite the estimation of $M_{t}$ as follows:
\begin{equation}
\footnotesize
\hat M_{t} = MF_{t} +  f_{_{N}}(f_{_{N}}(AHGCSP(\hat S_{t}) * MDR_{t}^{W}) * (I_{t} - IF_{t})
\label{equ:mdp}
\end{equation}
where $t \in [t'-1, \cdots, t'-Q+1]$. Note that the initial $\hat S_{t'-Q+1}=[\hat M_{t'-Q}, \cdots, \hat M_{t'-P}]=[MF_{t-Q}, \cdots, MF_{t-P}]$, i.e. output of prior estimator and $K=P-Q+1$.
Up to now, we have estimated full OD matrix sequence from $[t'-1, \cdots, t'-Q+1, t'-Q, \cdots, t'-P]$ by the prior estimator and AHGCSP based estimator. Afterward, it is fed into AHGCSP to obtain the final prediction $\hat M_{t}$.

\begin{table*}[htb]
	\centering
	\caption{Model Performance of Different Methods on Different Datasets}
	\label{tab:Results}
	\begin{threeparttable} 
	\begin{tabular}{p{72pt}|p{48pt}|p{48pt}|p{48pt}|p{48pt}|p{48pt}|p{48pt}|p{48pt}|}
\toprule
\textbf{Datasets}&\multicolumn{3}{c|}{SZ05}&\multicolumn{3}{c}{SH04}\\ \cline{1-7}
\textbf{Metrics}	&	MAE	&	RMSE	&	WMAPE	&	MAE	&	RMSE	&	WMAPE\\ \cline{1-7}
HA& 1.0412±0.0000 & 3.5499±0.0000 & 0.7373±0.0000 & 0.9077±0.0000 & 2.2844±0.0000 & 0.7964±0.0000\\ \cline{1-7}
Ridge& 1.1839±0.0000 & 2.8113±0.0000 & 0.8019±0.0000& 0.9898±0.0000 & 2.0525±0.0000 & 0.8294±0.0000 \\ \cline{1-7}
TRMF & 0.8612±0.0212 & 2.7471±0.0238 & 0.6510±0.0014& 0.8368±0.0010 & 1.8189±0.0465 & 0.7152±0.0073 \\ \cline{1-7}
ANN& 0.7950±0.0102 & 1.8672±0.0442 & 0.5418±0.0068& 0.7145±0.0070 & 1.5848±0.0285 & 0.6015±0.0058 \\ \cline{1-7}
FC-LSTM& 0.7802±0.0024 & 1.8031±0.0053 & 0.5309±0.0016& 0.6890±0.0048 & 1.4556±0.0037 & 0.5811±0.0040 \\ \cline{1-7}
ConvLSTM& 0.7953±0.0101 & 1.8213±0.0174 & 0.5417±0.0068& 0.7135±0.0033 & 1.5112±0.0035 & 0.6007±0.0027 \\ \cline{1-7}
GCN& 0.7848±0.0150 & 1.8463±0.0597 & 0.5350±0.0101& 0.6949±0.0068 & 1.4911±0.0252 & 0.5852±0.0056 \\ \cline{1-7}
CASCNN & 0.7710±0.0015 & 1.7424±0.0022 & 0.5218±0.0035& 0.6855±0.0037 & 1.4326±0.0051 & 0.5792±0.0020 \\ \cline{1-7}
GEML& 0.7660±0.0030 & 1.6802±0.0039 & 0.5157±0.0020& 0.6837±0.0046 & 1.4166±0.0029 & 0.5701±0.0038 \\ \cline{1-7}
C-AHGCSP& \textbf{0.7127±0.0031} & \textbf{1.6133±0.0087} & \textbf{0.4767±0.0019}& \textbf{0.6470±0.0019} & \textbf{1.3904±0.0108} & \textbf{0.5356±0.0015} \\ \cline{1-7}
\bottomrule
\end{tabular}
\begin{tablenotes} 
\item Note: The model performance measured by various metrics above is tested on the same test dataset. To obtain more convincing and stable results, all the experiments are repeated ten times. The mean and the standard error of each metric are utilized to represent the final results. HA does not require training. It is only repeated once and the standard error of its metrics is zero. In addition, when the parameters of Ridge is fixed and the train dataset, test dataset are fixed, the results of the repeated experiment of Ridge are the same.
\end{tablenotes} 
\end{threeparttable} 
\end{table*}

\section{Experiments}
We conduct a large amount of experiments on two real-world subway datasets to find answers to the following research questions:

RQ1: Does our model have a better performance compared with other baselines on different metro network datasets?

RQ2: Does each component in our model help to improve the model performance?


RQ3: What about the main hyper parameters sensitivity of the model?

RQ4:What's the effectiveness of the proposed data completion method?

\subsection{Experiment Settings}

\begin{table*}[htbp]
	\centering
	 \begin{threeparttable} 
 	\caption{The Details of Different Datasets with 15 Minutes Granularity}
 	\label{tab:data}
 \begin{tabular}{ccccccccccccc}
  \toprule
  \textbf{Datasets}&Date&Days &Stations&Records&Max&Mean&Median&Std&[0, 1)&[1, 5)&[5, 10)&[10,Max]\\
  \midrule
\textbf{Shenzhen}&20140501-20140530&30&118&19.3&1388&1.59&0.0&5.7&62.1\%&29.0\%&5.4\%&3.5\%\\
 \midrule
 \textbf{Shanghai}&20150401-20150430&30&133&11.2&334&1.26&0.0&3.8&63.2\%&30.1\%&4.4\%&2.3\%\\
 \cline{1-13}
 	\end{tabular}
\end{threeparttable} 
\end{table*}

\subsubsection{\textbf{Datasets}}
Extensive experiments are conducted in two real-world subway datasets from China, i.e. Shenzhen (SZ), Shanghai (SH). Each dataset has million trip records by the automatic fare collection systems (AFCs). The datasets have been preprocessed to remove the passenger privacy information, complying with the security and privacy policies. Each trip is composed of five elements, i.e. the ID number of smart card, the ID of origin station, the entering time, the ID of destination station, the exiting time. Most metro lines have different operation time. To be consistent, we do research on operation time period from 7:00 to 23:00. The trips are aggregated at a time granularity of 15 minutes, thus generating $\text{T}=64$ time slots in one day. Z-score is leveraged to normalize the datasets. As to the data split, 70\% of each dataset is divided into train data, 10\% as validation data and 20\% as test data. The details of datasets and some statistical information of the OD values are elaborated in Table \ref{tab:data}.

\subsubsection{\textbf{Baselines}} We choose various kinds of approaches as our baselines, including the traditional methods, deep learning methods and the state-of-the-art method.

\textbf{(1) HA} (Historical Average) is the most widely used model to predict time sequential data. It averages the historical value of an OD pair to predict its future value at the next time slot. HA has no trainable parameters and training process. We directly evaluate HA on the test dataset.

\textbf{(2) Ridge} is a linear regression method with L2 regularizer which tends to assign weights to different input features evenly. 

\textbf{(3) TRMF}\cite{Yu2016TemporalRM} (Temporal Regularized Matrix Factorization) is  a matrix factorization method with autoregression (AR) processes on each temporal factor.

\textbf{(4) ANN} (Artificial Neural Network) is a simple neural network which can extract linearity and non-linearity of data in some extent. Our ANN has three layers with units $[256, 256, N]$ where $N$ is the number of stations. We choose ReLU as the activation function of all layers.

 \textbf{(5) FC-LSTM} (Fully connected-Long Short Term Memory Network) is a classical neural network for time series forecasting which can extract the long term temporal dependency in time series data. Our FC-LSTM has two hidden layers with 256 units and an output layer has N units with ReLU activation function. FC-LSTM is adopted to predict the destination distribution of the target station with its historical destination distributions as input. 
 
\textbf{(6) ConvLSTM} \cite{shi2015convolutional} is a variant of FC-LSTM by replacing the fully connection with the convolution operation. Compare with FC-LSTM which can only extract the long-term temporal pattern from time-series data, ConvLSTM can capture the temporal dependency as well as grid-based spatial dependency. The setting of our ConvLSTM is the same as that in previous works \cite{zhang2021short}, which has three layers with 8, 8, 1 filters respectively and 3 x 3 kernel size of all the filters.

\textbf{(7) GCN}\cite{kipf2016semi} (Graph Convolution Network)  can extract the spatial dependency through aggregating the features from neighbors in a traffic graph. 
 
 \textbf{(8) CASCNN}\cite{zhang2021short} (Channel-wise Attentive Split Convolutional Neural Network) is a CNN based deep learning model to predict the short-term OD matrix in metro system. It utilized split CNN and channel-wise attention mechanism to process the historical OD data matrices at previous days and it designed an inflow/outflow-gated mechanism to merge the historical OD flow information with real-time inflow/outflow information.
 
\textbf{(9) GEML}\cite{wang2019origin} (Grid-Embedding based Multi-task Learning) aims to solve the rail-hailing Origin-Destination matrix prediction with previous origin-destination matrices. Specifically, GEML utilizes GCN to capture the spatial dependency from the geographical neighborhood and semantic neighborhood of the target station and it adopts LSTM to capture temporal attributes of the passenger destination distribution. GEML has three predictive objectives, i.e. inflow, outflow and OD matrix prediction. To be consistent with other baselines and our model, we only keep its OD matrix prediction objective.

\subsubsection{\textbf{Evaluation Metrics}}
All model performances are evaluated with three wildly applied metrics in previous works \cite{wang2019origin, zhang2021short, cheng2021real}, i.e. MAE (Mean Absolute Error), RMSE (Root Mean Square Error) and WMAPE (Weighted Mean Absolute Percentage).
\begin{equation}
\begin{aligned}
MAE&=\frac{1}{N \times N} \sum_{i=1}^{N}\sum_{j=1}^{N} \lvert m_{t}^{ij}-\hat{m}_{t}^{ij} \lvert \\
RMSE&=\sqrt{\frac{1}{N \times N} \sum_{i=1}^{N}\sum_{j=1}^{N}(m_{t}^{ij}-\hat{m}_{t}^{ij})^2}\\
WMAPE&=\frac{\sum_{i=1}^{N}\sum_{j=1}^{N} \lvert m_{t}^{ij}-\hat{m}_{t}^{ij} \lvert }{\sum_{i=1}^{N}\sum_{j=1}^{N}m_{t}^{ij}}
\end{aligned}
\end{equation}

MAE and RMSE can only be compared among the same dataset and they are sensitive to data scale. WMAPE can be adopted to evaluate model performance on different datasets. WMAPE is data-scale independent and can avoid zero-division and over-skewing confronting by MAPE. WMAPE is used to measure prediction of the whole OD matrix instead of OD pair level.

\subsubsection{\textbf{Parameter Settings}}
To all the models, the input time slot $P$ is set as 8. The trainable parameters of all deep learning models are tuned on validation set. As our model, the input time slot $Q$ in prior estimator is set as 5 according to the data incomplete analysis.  The best units of $f_{key}$, $f_{query}$, $\bar f_{key}$, $\bar f_{query}$ in our real-time global dynamicity module is tuned as $[16, 16, 16, 16]$ in both SH and SZ. The best units of our graph convolution are tuned as $[512, 512]$ in SH and $[128, 128]$ in SZ . The best units of our LSTM are tuned as $[256, 256]$ both in SH and SZ. Note that AHGCSP and AHGCSP based estimator in our model do not share the same trainable parameters but for simplicity, they share the same hyper parameters. All the parameters are optimized to minimize the loss function in Equation \ref{lab:equ1} by Adam optimizer \cite{kingma2014adam}.  The batch size for both datasets is set as 32. Learning rate decay strategy is adopted to get the best result of all the deep learning models. The initial learning rate of our model is set as 0.01. The decay ratio of learning rate is set is 0.9. The learning rate will decay if the validation result does not improve within 15 epochs. The minimum learning rate is 2.0e-06. The maximum number of epochs is set as 1000. Python and TensorFlow are utilized for the code and experiments are running on a GPU (32GB) with TESLA V100.

\begin{figure*}[htp]
	\centering
	\subfigure[MAE of C-AHGCSP]{
		\includegraphics[width=0.31\textwidth,height=0.12\textheight]{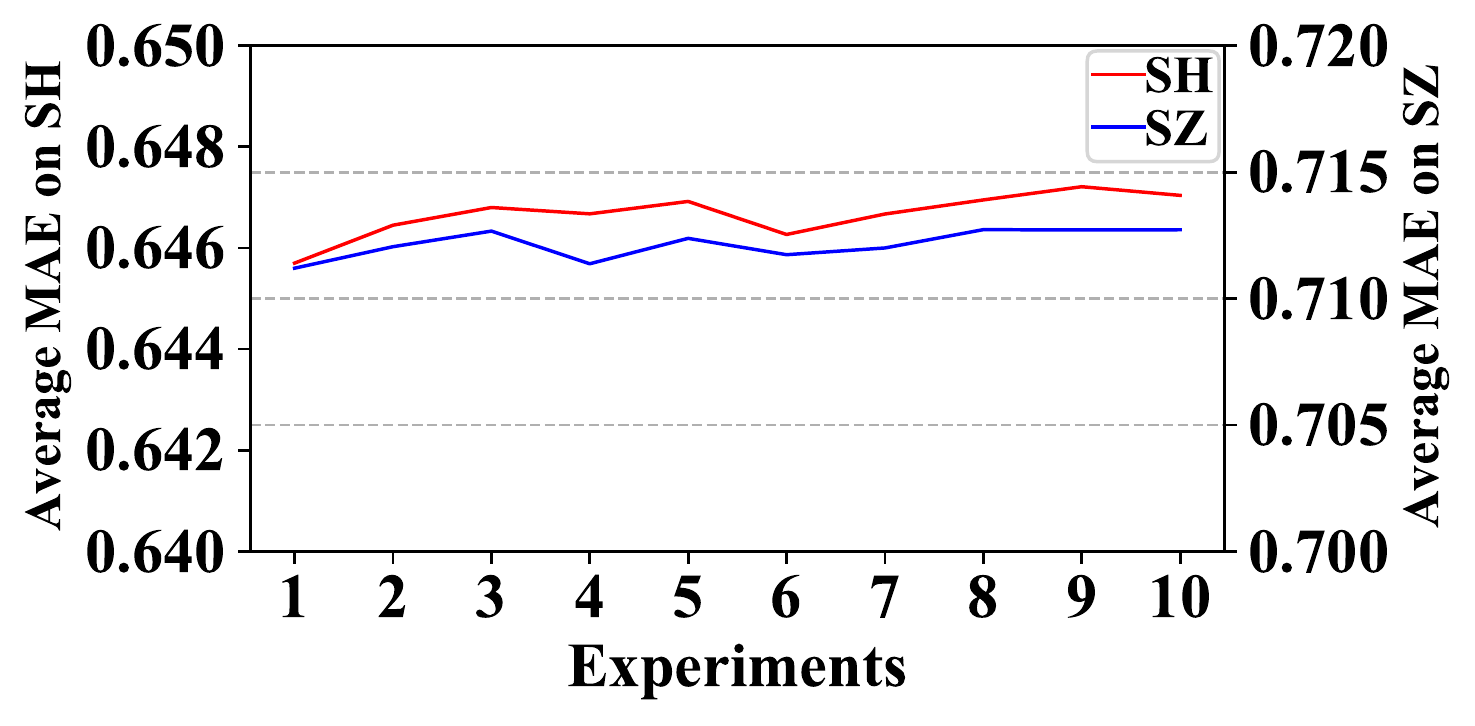}
	}
	\subfigure[RMSE of C-AHGCSP]{
		\includegraphics[width=0.31\textwidth,height=0.12\textheight]{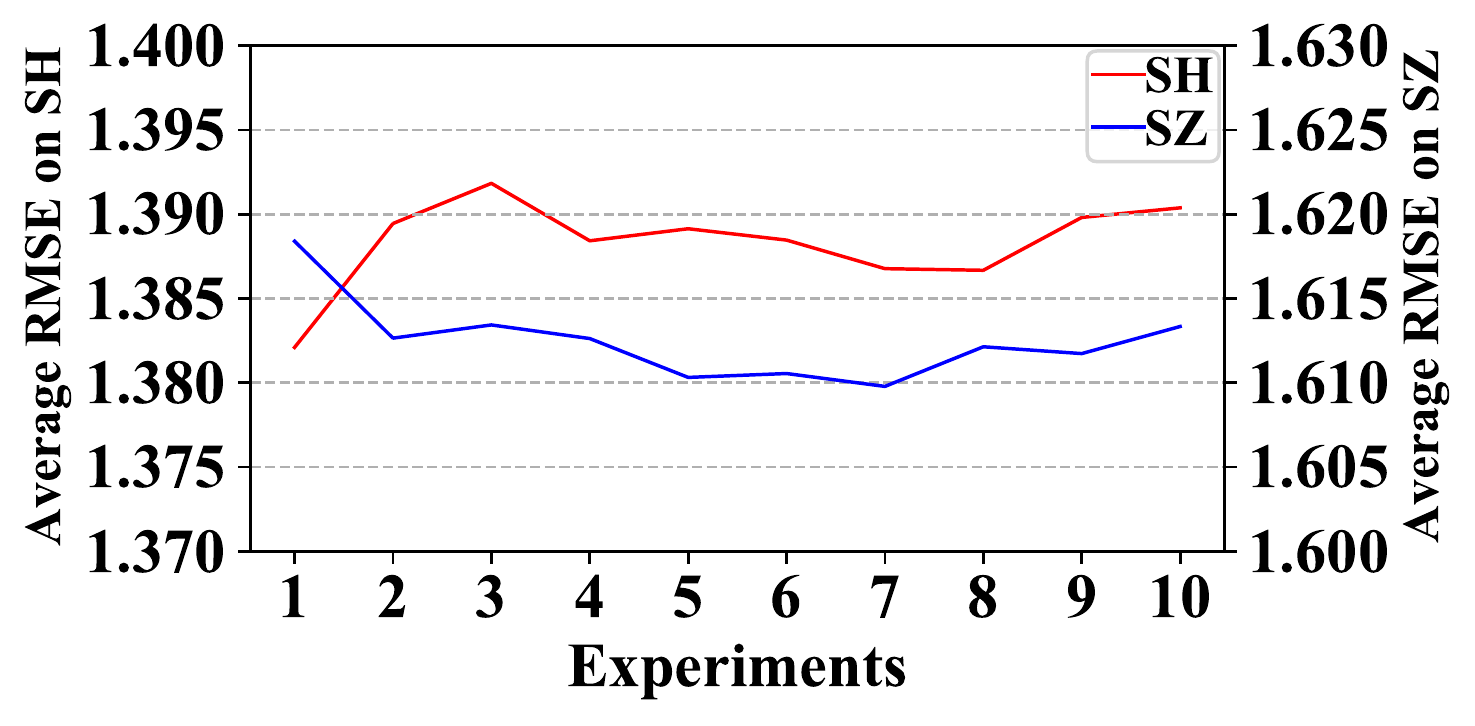}
	}
	\subfigure[WMAPE of C-AHGCSP]{
		\includegraphics[width=0.31\textwidth, height=0.12\textheight]{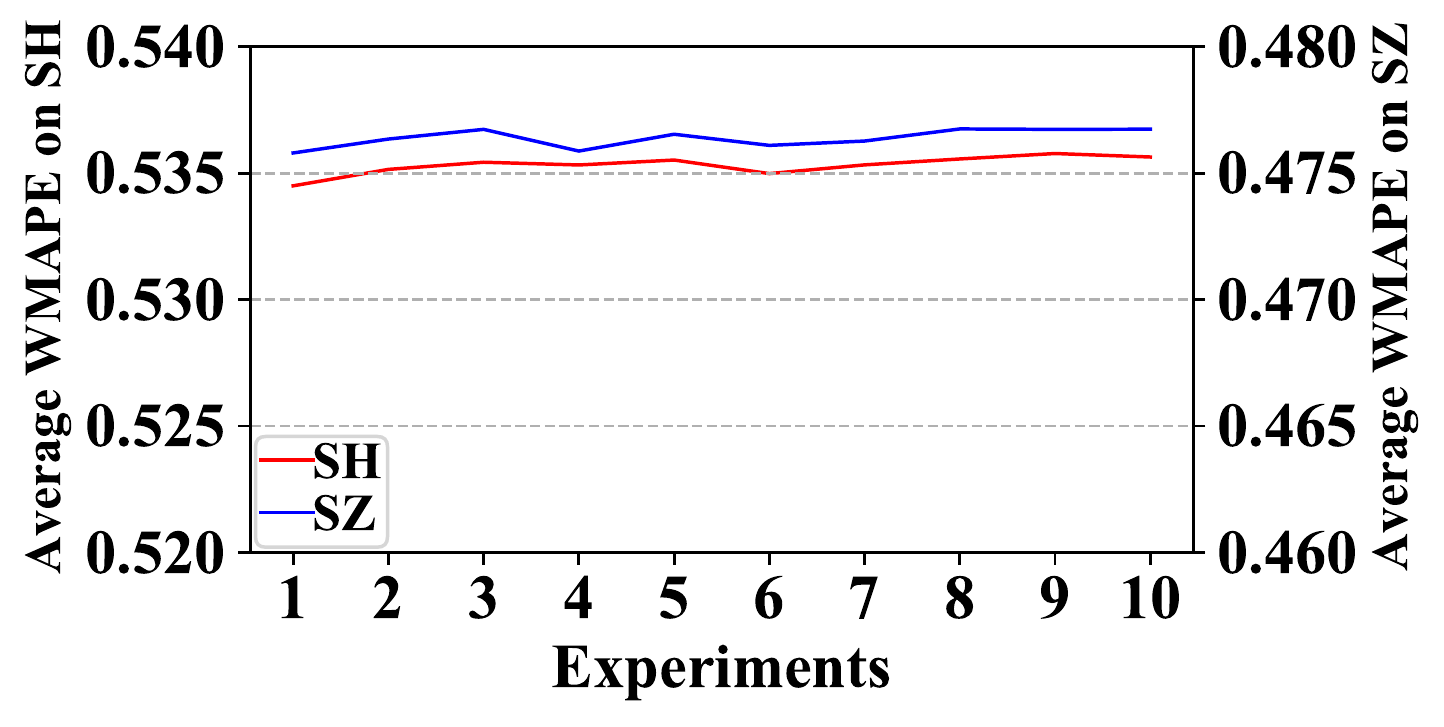}
	}
        \caption{We analyze the experimental stability on two datasets. Take Figure (a) for example,  as the number of the experiments increases, the average value of MAE fluctuates in a very small scale, showing the stability of the model performance.}
	\label{fig:sta}
\end{figure*}

\begin{figure*} [!htb]
 \centering
 \subfigure[OD Pair 1 (SZ)]{
  \includegraphics[width=0.28\textwidth,height=0.13\textheight]{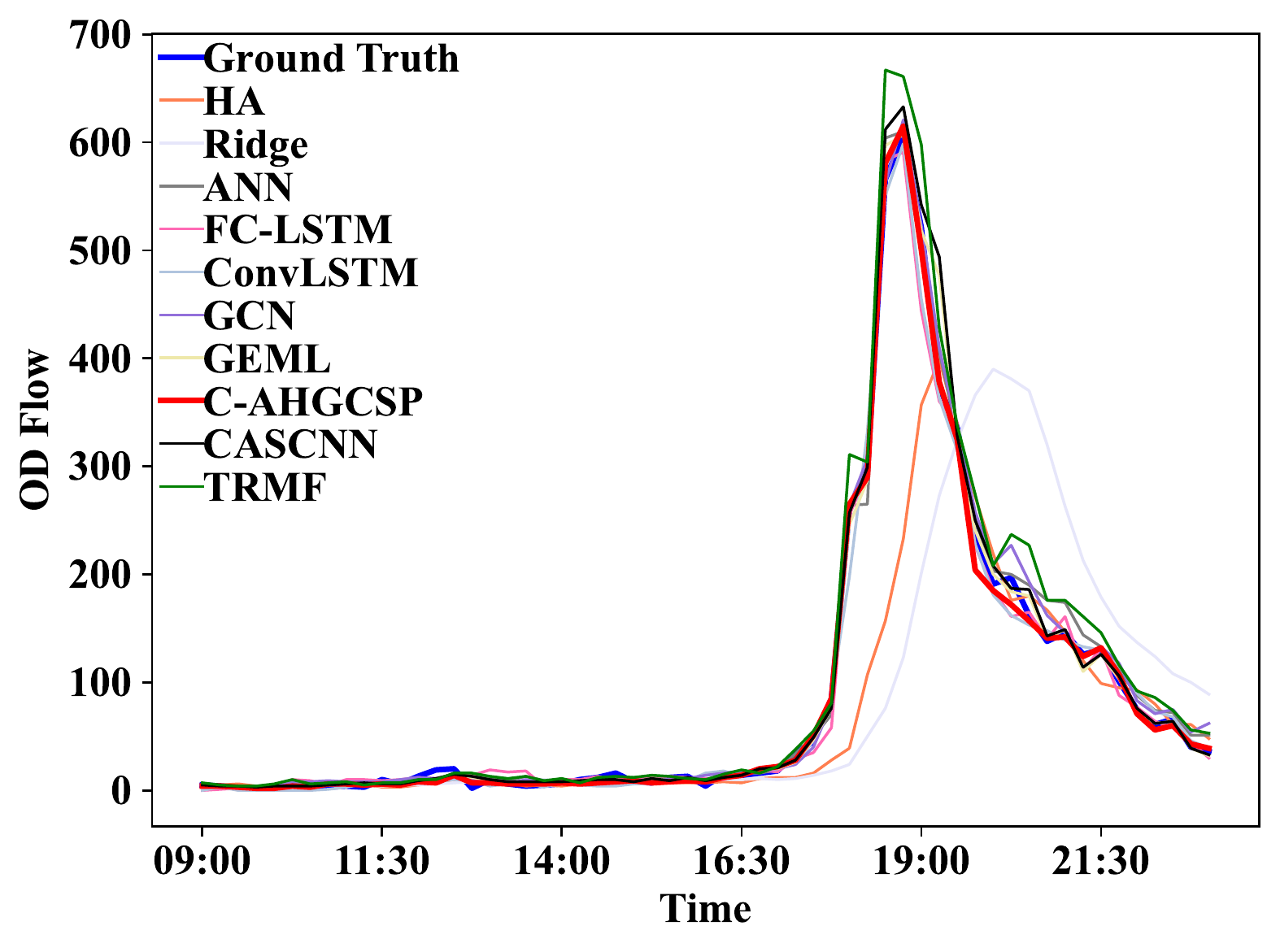}
 }
 \subfigure[OD Pair 2 (SZ)]{
  \includegraphics[width=0.28\textwidth,height=0.13\textheight]{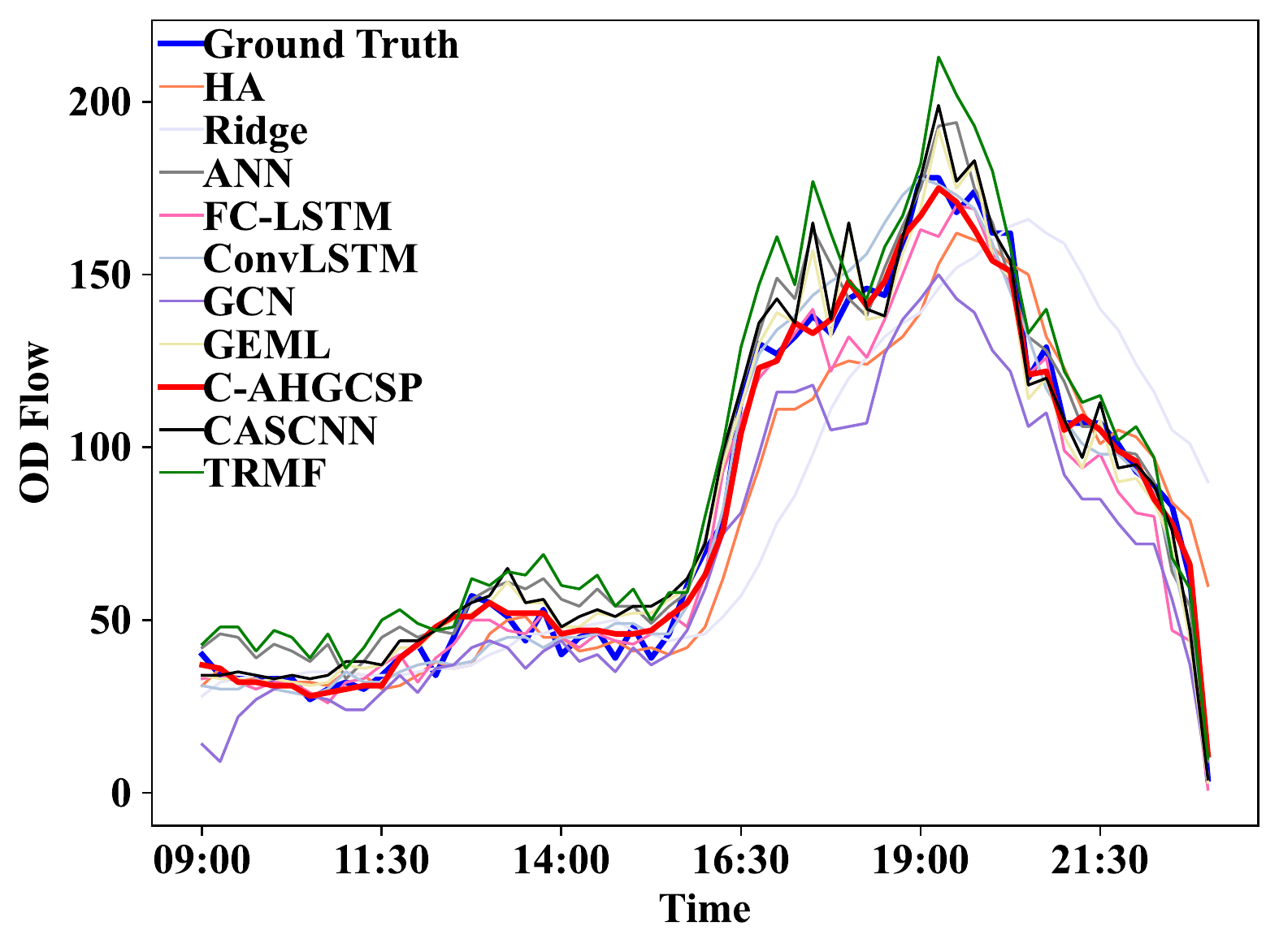}
 }
 \subfigure[OD Pair 3 (SZ)]{
  \includegraphics[width=0.28\textwidth,height=0.13\textheight]{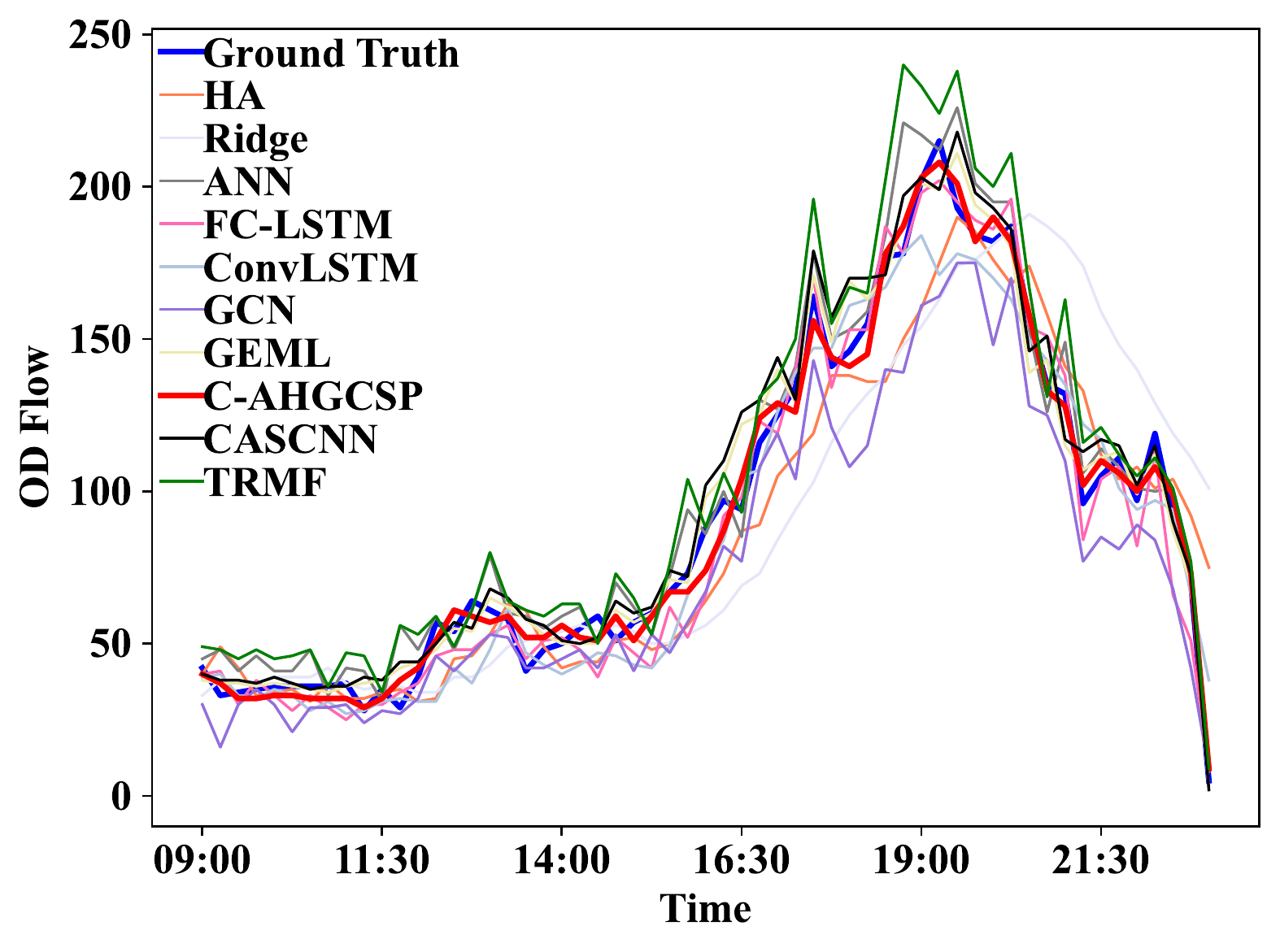}
 }
\subfigure[OD Pair 1 (SH)]{
 \includegraphics[width=0.28\textwidth,height=0.13\textheight]{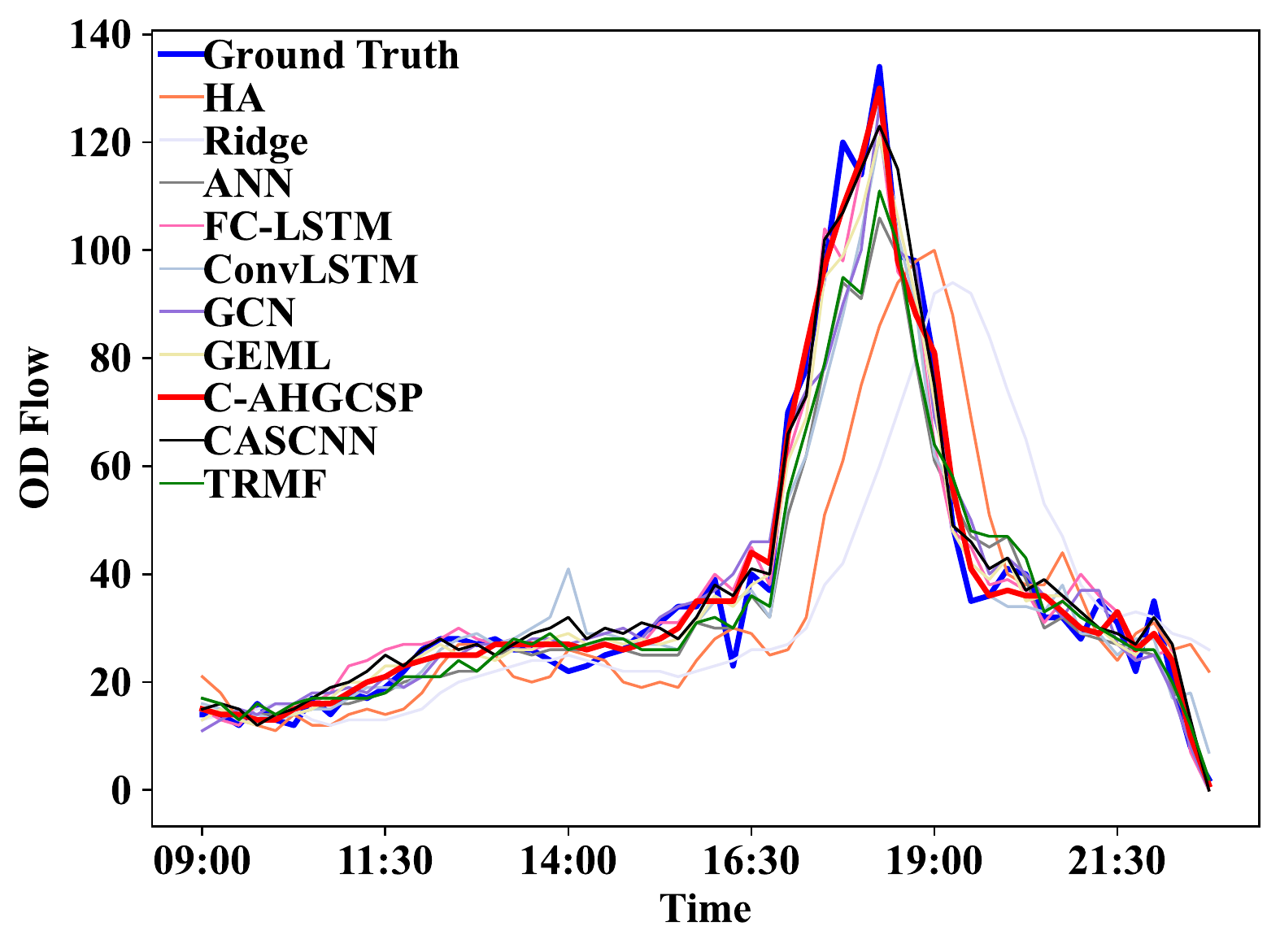}
}
\subfigure[OD Pair 2 (SH)]{
 \includegraphics[width=0.28\textwidth,height=0.13\textheight]{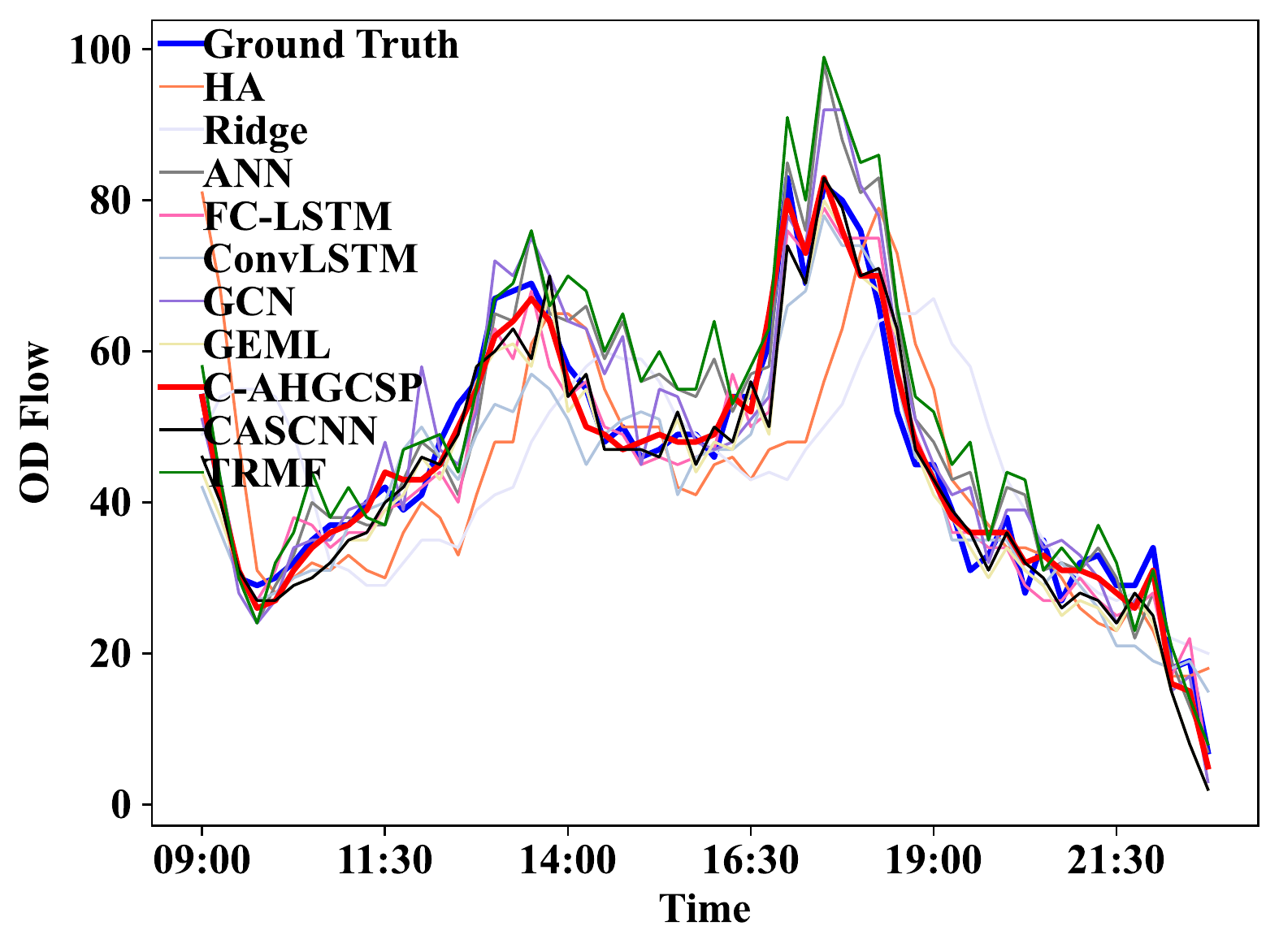}
}
\subfigure[OD Pair 3 (SH)]{
 \includegraphics[width=0.28\textwidth,height=0.13\textheight]{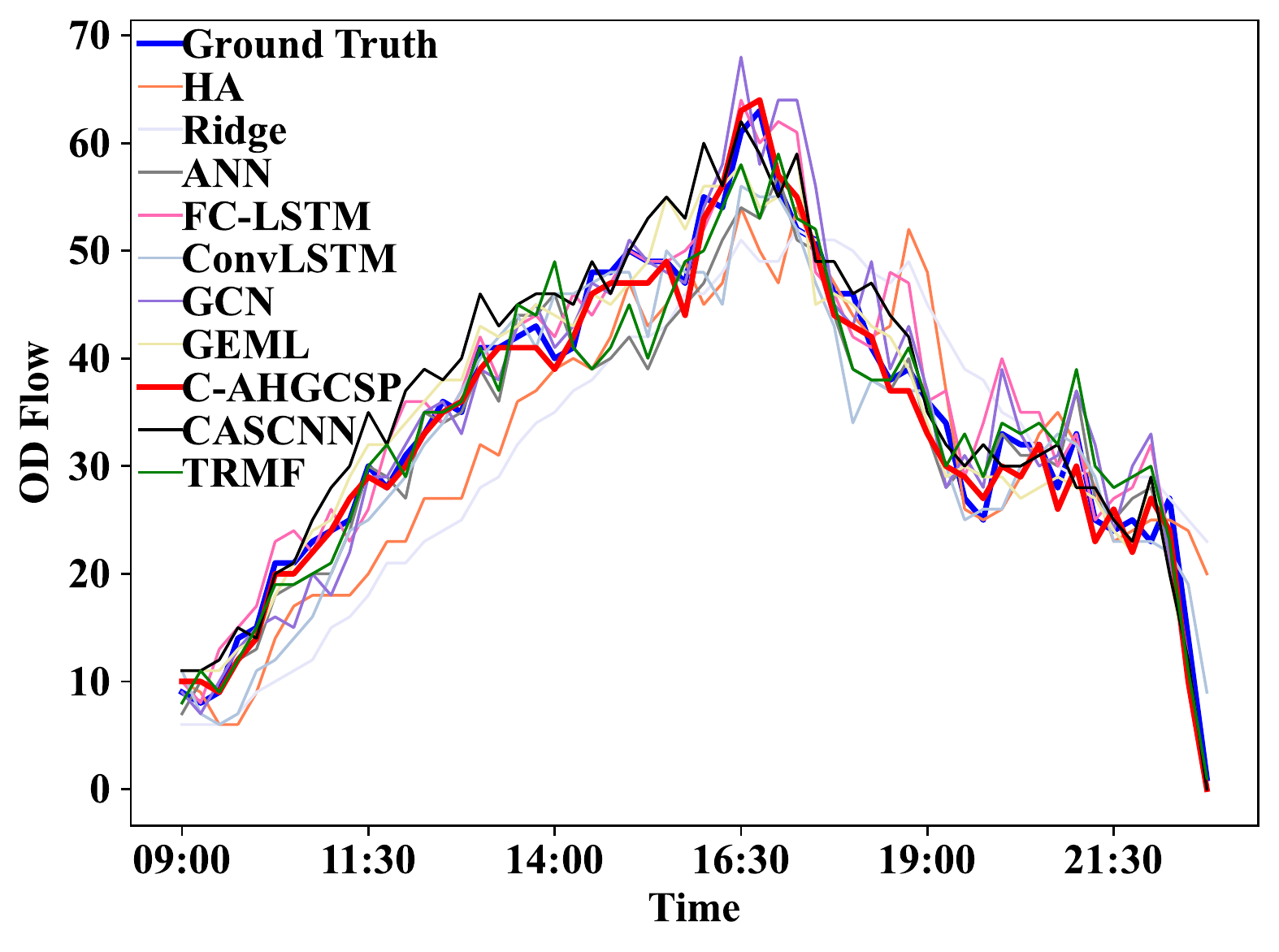}
}
\caption{We pick three OD pairs with the largest OD flow in a randomly chosen day and visualize the ground truth and prediction of all models on both Shenzhen and Shanghai Datasets. The blue line is the ground truth of OD value and the red line is the prediction of our model. It can be observed from the figures above that our model fits the ground truth better than other baselines.}
 \label{fig:vis}
\end{figure*}

\subsection{Experimental Analysis}
The experiment results in Table \ref{tab:Results} are obtained from the test data and each experiment is repeated ten times to get more convincing results. The best results are highlighted in bold. As we can see from the Table, the standard error of each experiment result is quite small, indicating that the repeated experiments are stable. We visualize the average results of ten experiments on three metrics of our model in Figure \ref{fig:sta}.

\subsubsection{\textbf{Performance Comparison}}
Table \ref{tab:Results} shows all models performances on two metro datasets. Our model C-AHGCSP achieves the best performance at all metrics on all the datasets. Specifically, our model improves the prediction accuracy by \textbf{3.9\%, 3.45\%}  (WMAPE) on Shenzhen and Shanghai datasets compared with GEML, a model with suboptimal performance. It can be seen that traditional methods like HA, Ridge, TRMF do not perform well and have at least \textbf{10\%}WMAPE difference compared with deep learning models because they lack the capacity to deal with a large amount of data and to extract the non-linearity in the OD data. Deep learning methods generally have better performances than conventional approaches for that they are more suitable to enormous data and non-linearity. ANN has the worst performance perhaps because it can't extract spatiotemporal correlations. FC-LSTM can capture the long-term dependency in OD flows and its WMAPE is slightly better than ANN, namely about 1\% improvement on Shenzhen dataset and about 2\% improvement in Shanghai dataset. ConvLSTM  can also extract the temporal pattern in OD data. However, it also considers the OD matrices as images to model the spatial dependency. From the experimental results, it can be observed that such image-based spatial dependency extraction is inappropriate and it worsens the performance of ConvLSTM compared with LSTM. GCN can not learn temporal correlation while it models the metro system as graph, closer to its nature, which is verified by the fact that GCN has a better performance than ConvLSTM. The performances of LSTM and GCN are quite close and it infers that both temporal property and spatial property exist in OD flow pattern.
CASCNN is better than the simple deep learning methods, e.g. GCN, FC-LSTM. It made full advantage of the real-time inflow/outflow information. However, it compacts OD matrices  at previous days to obtain dense information by CNN and it does not pay attention to the temporal pattern in OD flow and it also ignores that graph nature topology of metro system. GEML has a better performance than CASCNN.
GEML can capture spatiotemporal correlations in the traffic network based on graph. However, it is fed with the incomplete real-time OD data. What's worse, it does not take the functional correlation and real-time global dynamicity into consideration. Compared with GEML, our model proposes to complete the recent OD matrices to obtain more real-time passenger flow information and we simultaneously consider three spatial dependencies (i.e. both local and global, static and dynamic spatial properties) in OD data based on a subway graph. Consequently, C-AHGCSP achieves the best performance among all the methods. The prediction of all models and the ground truth is visualized on Figure \ref{fig:vis}.

\begin{table}[htbp]
	\caption{Model Ablation In Shenzhen Dataset}
	\label{tab:abl}
	\centering	
	\begin{tabular}{cccc}
		\toprule
		\textbf{Modules} &\textbf{MAE}&\textbf{RMSE}&\textbf{WMAPE}\\
		\midrule
		$\textbf{C-AHGCSP}_{\textbf{Geo}}$&0.7248$\pm$0.0027&1.6249$\pm$0.0054&0.4870$\pm$0.0012\\
		$\textbf{C-AHGCSP}_{\textbf{KL}}$&0.7355$\pm$0.0019&1.6363$\pm$0.0035&0.4895$\pm$0.0016\\
    		$\textbf{C-AHGCSP}_{\textbf{Dym}}$&0.7451$\pm$0.0021&1.6512$\pm$0.0066&0.4912$\pm$0.0020\\
		$\textbf{C-AHGCSP}_{\textbf{GCN}}$&0.7675$\pm$0.0027&1.6630$\pm$0.0091&0.4946$\pm$0.0017\\
		$\textbf{C-AHGCSP}_{\textbf{Com}}$&0.7703$\pm$0.0017&1.6701$\pm$0.0073&0.5049$\pm$0.0014\\
                  $\textbf{C-AHGCSP}$& \textbf{0.7127±0.0031} & \textbf{1.6133±0.0087} & \textbf{0.4767±0.0019}\\
		\cline{1-4}
	\end{tabular}
\end{table}

\subsubsection{\textbf{Model Component Analysis}}
\label{sub:com}
This section aims to test the contribution of each component of our model. To achieve this goal, we first remove each module of C-AHGCSP to obtain degraded variants as follows:

$\textbf{C-AHGCSP}_{\textbf{Com}}$: The data complete estimator is removed from our model.

$\textbf{C-AHGCSP}_{\textbf{Geo}}$: The Geographical Proximity is removed from AHGCSP module,  namely Equation \ref{equ:ada} becomes $\alpha_{i j}^{t} = w_{d} \alpha_{i j}^{dt} + w_{f}\alpha_{i j}^{f}$. Note that data complete estimator does not change and Geographical Proximity is not removed from AHGCSP based  estimator.

$\textbf{C-AHGCSP}_{\textbf{KL}}$: The KL Divergence based Functional Correlation is removed from AHGCSP module, namely Equation \ref{equ:ada} becomes $\alpha_{i j}^{t} = w_{d} \alpha_{i j}^{dt} + w_{g}\alpha_{i j}^{g}$. Note that the data complete estimator does not change and Functional Correlation is not removed from AHGCSP based estimator.

$\textbf{C-AHGCSP}_{\textbf{Dym}}$: The Real-Time Global Dynamicity is removed from AHGCSP module, namely Equation \ref{equ:ada} becomes $\alpha_{i j}^{t} = w_{f}\alpha_{i j}^{f} + w_{g}\alpha_{i j}^{g}$. Note that the data complete estimator does not change and Real-Time Global Dynamicity is not removed from AHGCSP estimator.

$\textbf{C-AHGCSP}_{\textbf{GCN}}$: All three spatial dependency and the graph convolution are removed from AHGCSP module. The input data is transformed by the the data complete estimator, then it is fed into LSTM directly without spatial dependency extraction.

Ablation tests are conducted on these variants to demonstrate the contribution of each part in our model. Due to the space limitation, Table \ref{tab:abl} only shows the ablation results on Shenzhen dataset. We train all the variants by repeating the experiments 10 times to get stable results. All the model performance are obtained on the test dataset. It can be seen from Table \ref{tab:abl} that no matter which module is removed, the model performance becomes worse, showing that all modules are necessary to improve prediction accuracy. However, their contributions are different. The data completion module contributes the most, i.e. about \textbf{2.8\%}, which proves the effectiveness of the Full OD Matrix Reconstruction module. Without this module, our model performance degrades to 0.5049 in WMAPE, which is only slightly better than the model performance of GEML with 0.5157 in WMAPE. The second largest contributor is the heterogenous spatial correlation based graph convolution module. If none of the spatial dependency is captured in our model, its performance will decrease to 0.4946 in WMAPE, i.e. \textbf{1.8\%} improvement to our model. It proves that the spatial dependency extraction is important for improving OD prediction accuracy. As for the three kinds of spatial dependency, Real-Time Global Dynamicity contributes the most to the prediction, i.e. 1.45\% in WMAPE, while KL Divergence based Functional Correlation contributes 1.28\% improvement in WMAPE and Geographical Proximity contributes 1.03\% in WMAPE improvement. It indicates that real-time global dynamicity of passenger mobility has a larger impact on OD flow prediction than the other two spatial dependency. But all spatial dependency can improve the model performance and they capture different passenger mobility patterns from different spatial perspectives.

\begin{table}[htbp]
	\centering
	\tiny
	 \begin{threeparttable} 
         \caption{The Time Cost of Each Epoch of Deep Learning Models on SZ Dataset}
         \label{tab:tim}
	\begin{tabular}{cccccccc}
 	  \toprule 
	\textbf{Models} &\textbf{ANN}&\textbf{FC-LSTM}&\textbf{ConvLSTM}&\textbf{GCN}&\textbf{CASCNN}&\textbf{GEML}&\textbf{C-AHGCSP}\\
 	 \midrule
 	 \textbf{train (s)}&0.2&1.4&4.9&0.5&2.1&3.5&5.0\\
 	 \textbf{val (s)}&0.1&0.1&0.2&0.1&0.1&0.1&0.2\\
	 \textbf{test (s)}&0.1&0.2&0.7&0.1&0.1&0.5&0.8\\
 	 \textbf{total (s)}&0.4&1.7&5.8&0.6&2.4&4.1&6.0\\
	  \cline{1-8}
 	\end{tabular}
\end{threeparttable} 
\end{table}

\begin{table*}[htb]
	\centering
	\caption{Variant Model Performance Comparison}
	\label{tab:var}
	\begin{threeparttable} 
	\begin{tabular}{p{60pt}|p{48pt}|p{48pt}|p{48pt}|p{48pt}|p{48pt}|p{48pt}|p{48pt}|}
\toprule
\textbf{Datasets}&\multicolumn{3}{c|}{SZ05}&\multicolumn{3}{c}{SH04}\\ \cline{1-7}
\textbf{Metrics}	&	MAE	&	RMSE	&	WMAPE	&	MAE	&	RMSE	&	WMAPE\\ \cline{1-7}
C-ANN& 0.7855±0.0201 & 1.8267±0.0807  & 0.5307±0.0112& 0.7104±0.0086 & 1.5034±0.0192 & 0.5881±0.0121 \\ \cline{1-7}
C-FC-LSTM& 0.7415±0.0161 & 1.6529±0.0524  & 0.5134±0.0025& 0.6530±0.0183 & 1.3996±0.0175 & 0.5631±0.0051 \\ \cline{1-7}
C-ConvLSTM& 0.7950±0.0106 & 1.8593±0.0169  & 0.5387±0.0139& 0.7115±0.0081 & 1.5101±0.0288 & 0.5970±0.0137 \\ \cline{1-7}
C-GCN& 0.7737±0.0146 & 1.7809±0.0582 & 0.5232±0.0041& 0.6841±0.0045 & 1.4211±0.0036 & 0.5722±0.0046 \\ \cline{1-7}
C-GEML& 0.7311±0.0135 & 1.6440±0.0439 & 0.4961±0.0102& 0.6562±0.0031 & 1.4013±0.0034 & 0.5491±0.0125 \\ \cline{1-7}
C-AHGCSP& \textbf{0.7127±0.0031} & \textbf{1.6133±0.0087} & \textbf{0.4767±0.0019}& \textbf{0.6470±0.0019} & \textbf{1.3904±0.0108} & \textbf{0.5356±0.0015} \\ \cline{1-7}
AHGCSP based Estimator& 0.7001±0.0089 & 1.5912±0.0033 & 0.4601±0.0025& 0.6316±0.0036 & 1.3821±0.0092 & 0.5202±0.0046 \\ \cline{1-7}
\bottomrule
\end{tabular}
\begin{tablenotes} 
\item Note: Since CASCNN takes the OD matrices at previous days as input and they are complete so we do not evaluate the effectiveness of Data Completion on CASCNN. In addition, all models in above table except AHGCSP based Estimator predict the target OD matrix at $t'$ while 
AHGCSP based Estimator is utilized for completion of input OD matrix sequence at $[t'-1, \cdots, t'-Q+1]$.
\end{tablenotes} 
\end{threeparttable} 
\end{table*}

\subsubsection{\textbf{Model Efficiency Analysis}} This subsection aims to analyze the time cost of the deep learning baselines and our model. Due to the limited space, we only present the experiment efficiency on SZ dataset and the SH dataset shares similar conclusion. As shown in Table \ref{tab:tim}, ANN is the fastest model which only costs 0.2 second per epoch on the training stage due to its simplest network structure. Then comes GCN, it takes 0.5 seconds to finish an epoch. Compared with ANN, it has graph convolution operation besides FC layer.  FC-LSTM costs 1.4 seconds for that it has more trainable parameters, namely 939,638 while ANN has just 338,038 trainable parameters. CASCNN requires 2.1 seconds to be trained and it is mainly composed of convolution operation and attention mechanism. GEML requires 3.5 training seconds per epoch and it contains both LSTM and graph convolution operation. ConvLSTM needs more time than GEML for that it has both LSTM and convolution operation. The convolution operation is slower than graph convolution operation. Our model spends 5.0 seconds on the training stage per epoch which is nearly the same as ConvLSTM. Our model contains LSTM and graph convolution and the data completion process therefore it needs more time than other models. But its time cost is close to ConvLSTM while its performance is much better than ConvLSTM, namely about 7\% improvement in both datasets.

\subsubsection{\textbf{Hyper-parameter Analysis}}
\begin{figure}[h]
	\centering
	\subfigure[Shanghai Dataset]{
		\includegraphics[width=0.22\textwidth,height=0.11\textheight]{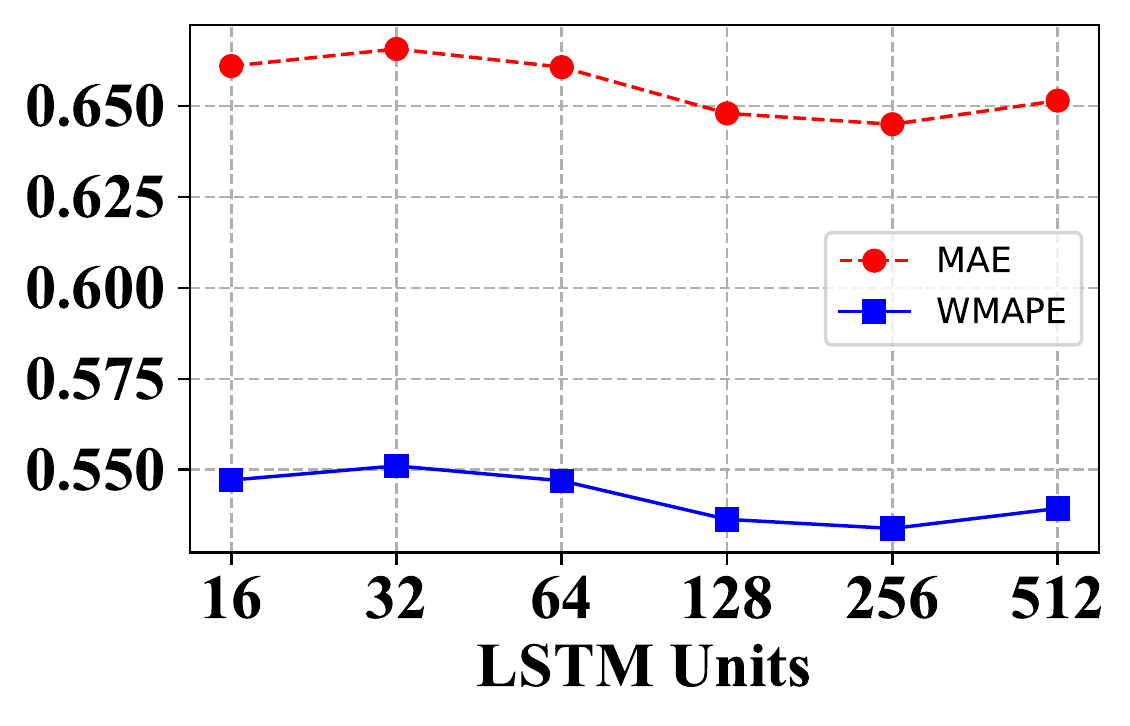}
	}
	\subfigure[Shenzhen Dataset]{
		\includegraphics[width=0.22\textwidth,height=0.11\textheight]{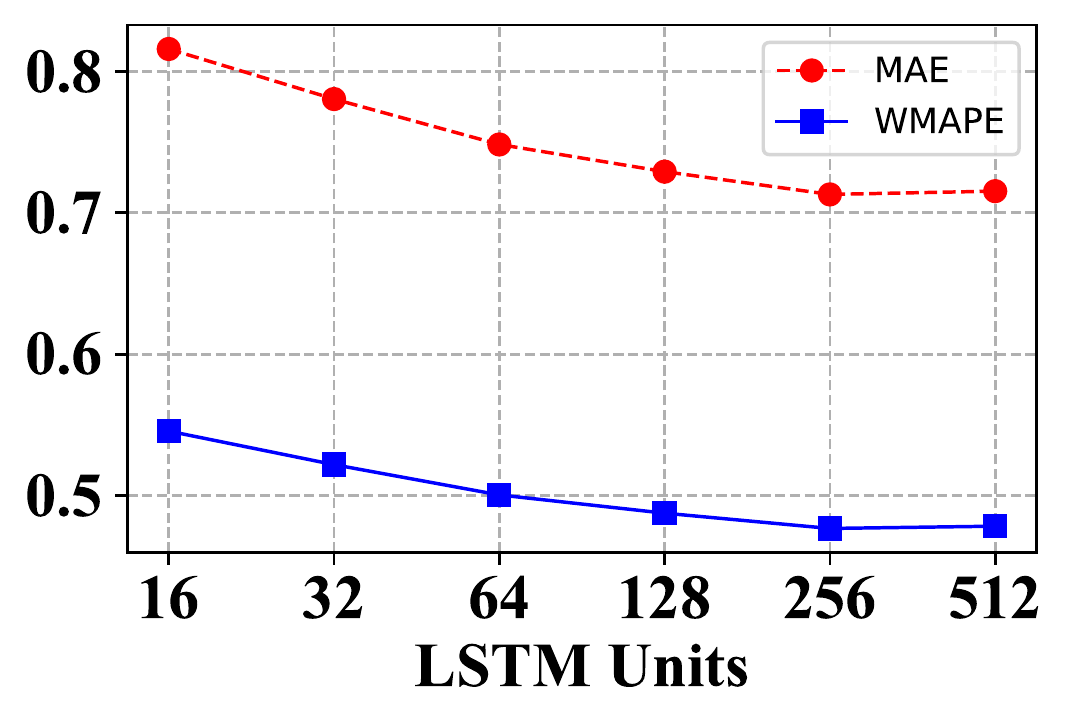}
	}
	\subfigure[Shanghai Dataset]{
		\includegraphics[width=0.22\textwidth,height=0.11\textheight]{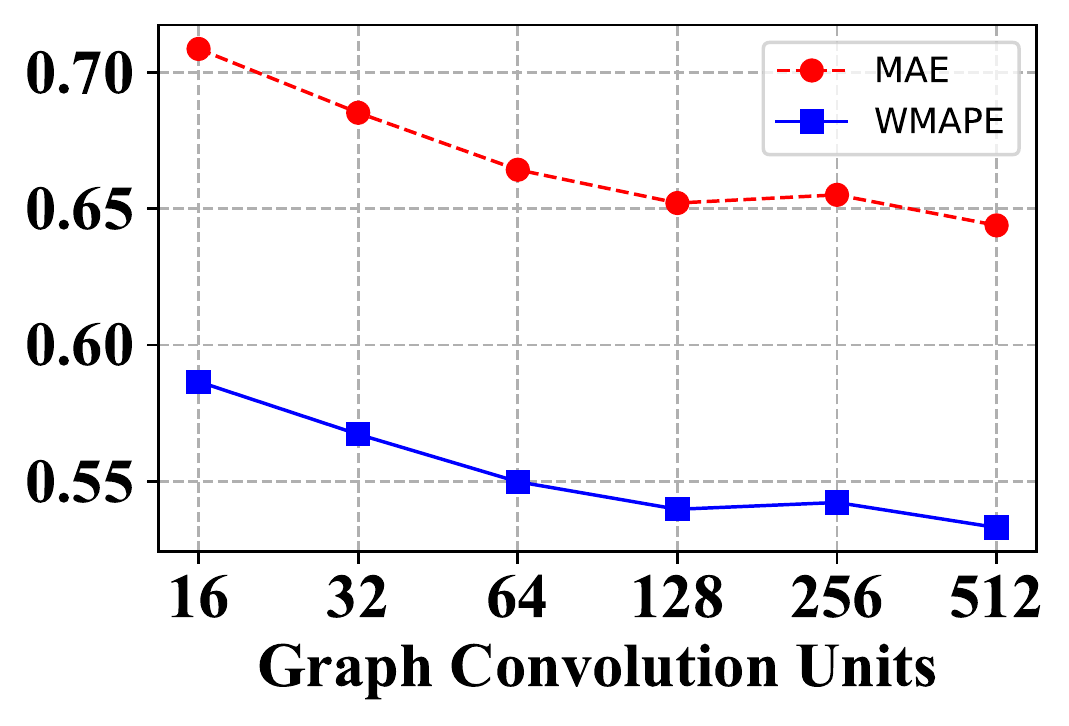}
	}
	\subfigure[Shenzhen Dataset]{
		\includegraphics[width=0.22\textwidth,height=0.11\textheight]{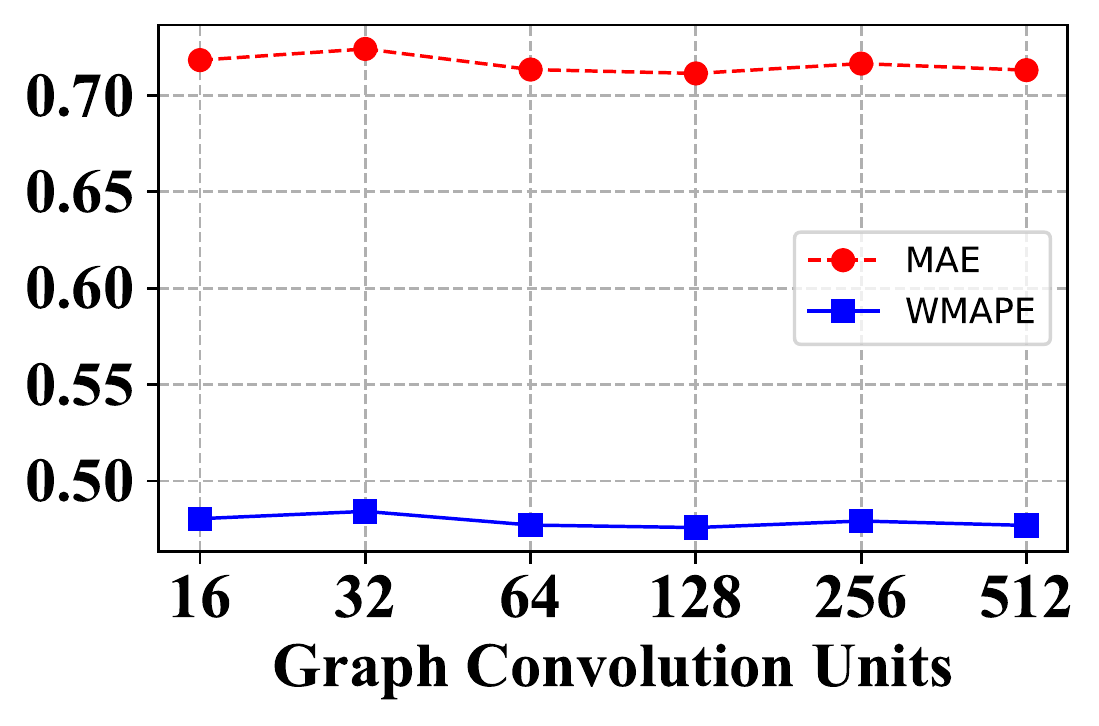}
	}
	\subfigure[Shanghai Dataset]{
		\includegraphics[width=0.22\textwidth,height=0.11\textheight]{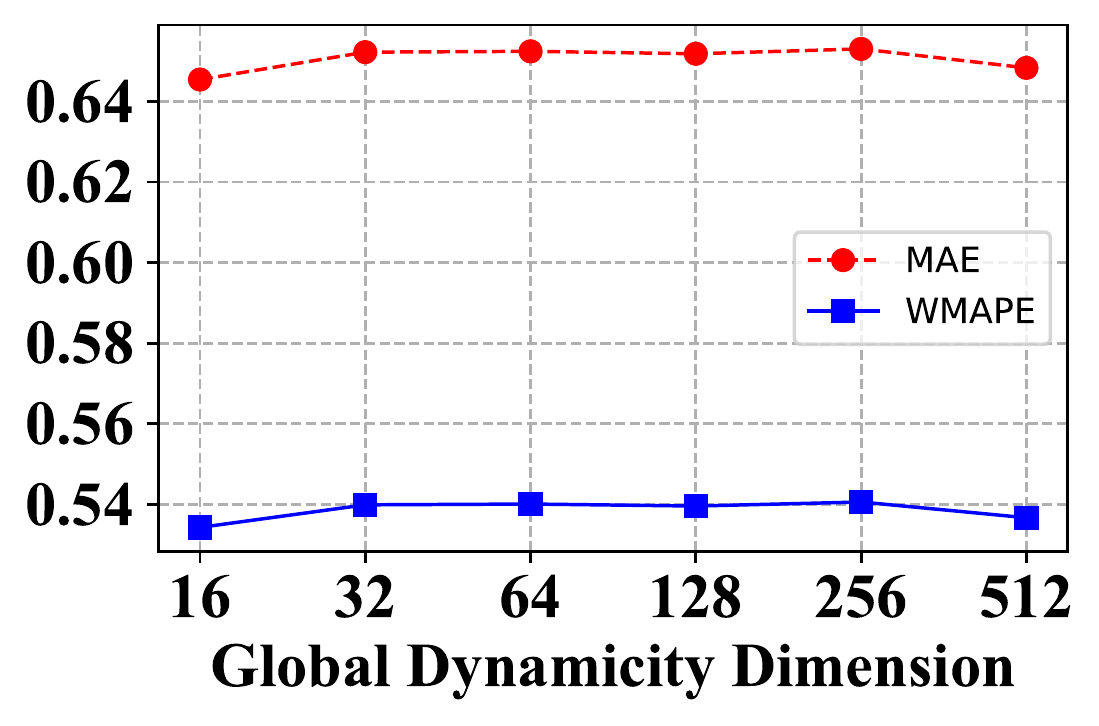}
	}
	\subfigure[Shenzhen Dataset]{
		\includegraphics[width=0.22\textwidth,height=0.11\textheight]{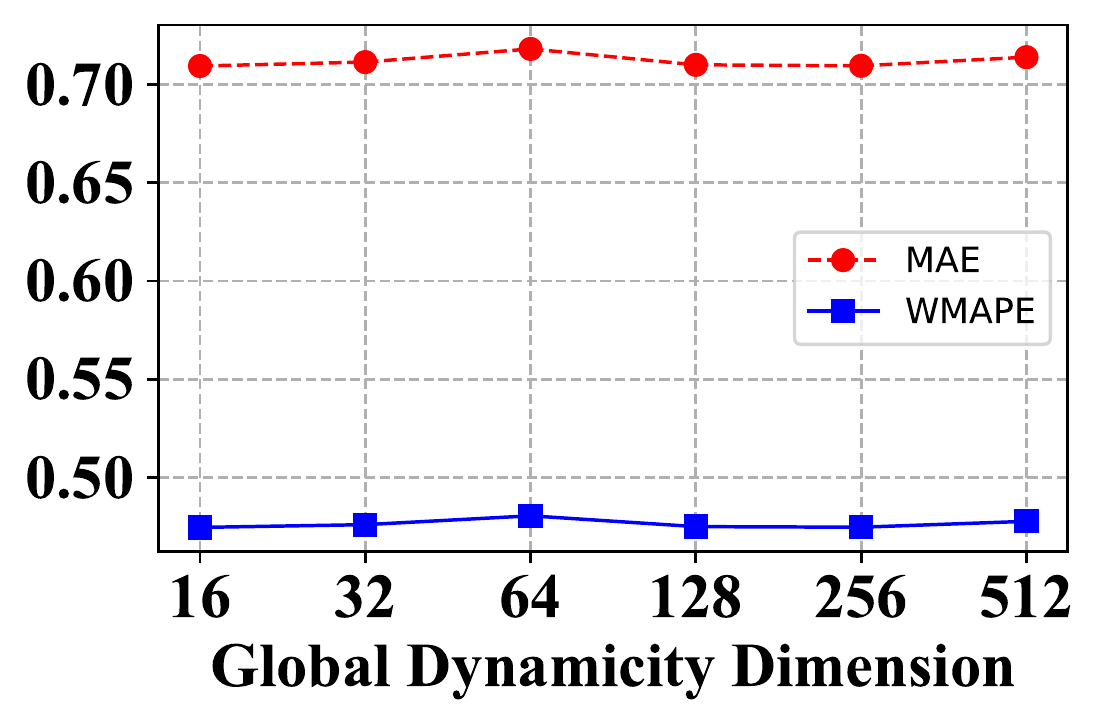}
	}
	\caption{Hyper Parameter Sensitivity Analysis on Two Datasets.}
	\label{fig:para}
\end{figure}

This section aims to analyze the hyper parameter sensitivity of our model on Shenzhen and Shanghai datasets. Other parameters are kept the same as default when the target parameter changes.

\textbf{Global Dynamicity Dimension} refers to the units of FC layer in the Key/Query part of Global Dynamicity extraction module.  The projection dimension of both key and value at data-driven correlation achieves best performance at 16 in both SH and SZ, much smaller than the number of metro stations. Such reduced dimension can help to concentrate the sparse OD flow information.

\textbf{Graph Convolution Units} refers to the units of graph convolution layer in the Graph Convolution based Heterogeneous Correlation Extraction module. When the graph convolution units is set at 512 on SH and 128 on SZ , the model achieves the best performance.

\textbf{LSTM Units} refers to the units of LSTM layer in the LSTM based Temporal Dependency Extraction module. We notice that as the number of LSTM units increases, the model has better performance. The best performance achieves at 256 at both datasets. After that it becomes worse, which might be caused by the overfitting phenomenon.

\begin{figure}[htb]
\centering
\includegraphics[width=0.480\textwidth, height=0.25\textheight]{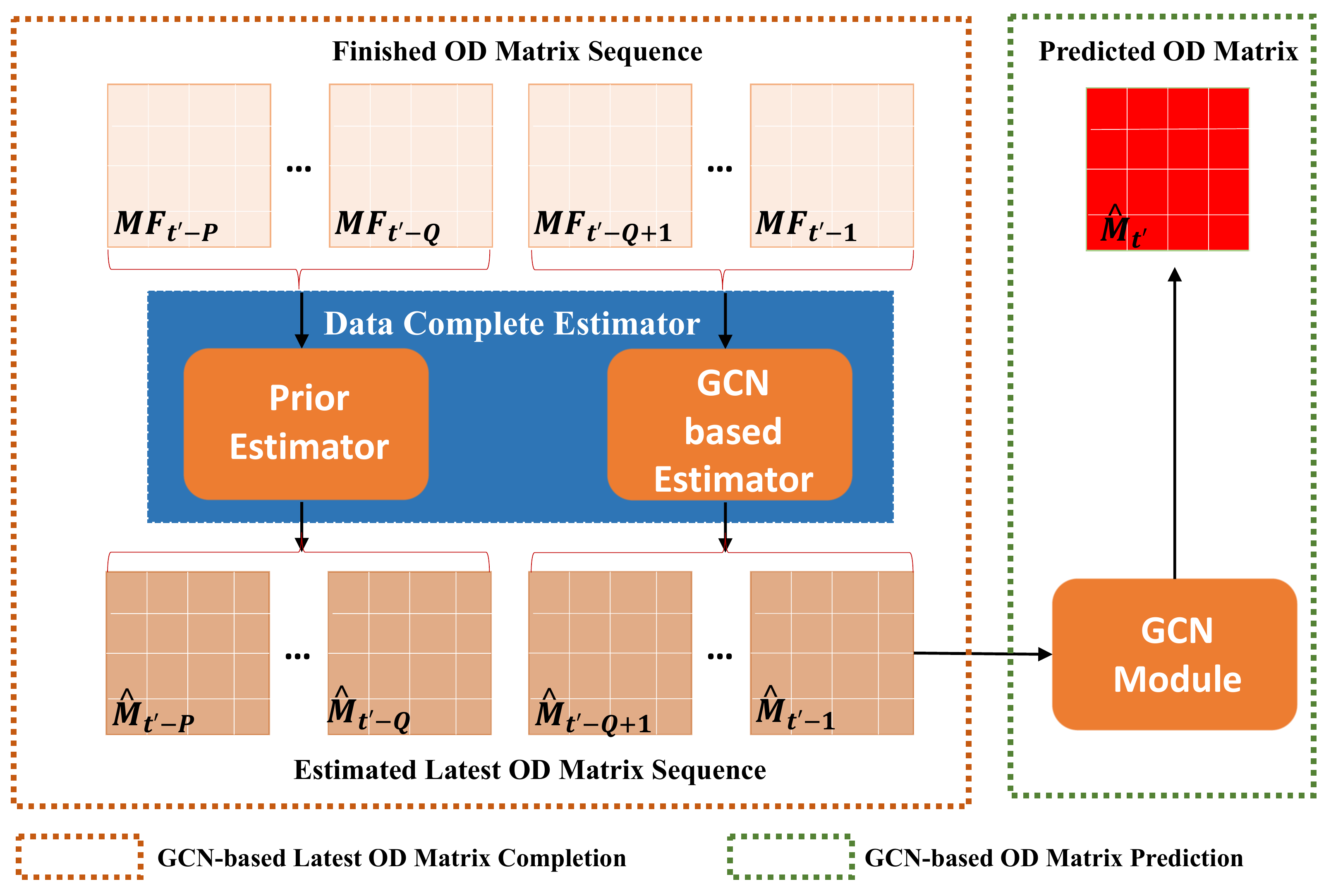}
\caption{The architecture of CGCN.}
\label{fig:cgcn}
\end{figure}

\subsubsection{\textbf{Effectiveness of Data Completion}}
As mentioned in Model Component Analysis section, the data completeness process can improve the C-AHGCSP performance for about 2.8\% in WMAPE in SZ datasets. Since we have the ground truth of the estimated OD matrix, we calculate the MAE, RMSE, WMAPE metrics based on the estimated OD matrices at $[t-Q+1, \cdots, t-1]$ and their ground truths (namely the full OD matrices), to compare the performance of AHGCSP based Estimator and the prediction of OD matrix at $t$. Note that the ground truth can only be collected after all the passengers reach their destinations. Table \ref{tab:var} shows the performance of AHGCSP based Estimator upon test data. We can observe that the performance of AHGCSP based Estimator is slightly better than the final time slot OD matrix prediction. This is perhaps due to the reason that prediction of OD matrix at $[t'-1, \cdots, t'-Q+1]$ incorporates more prior knowledge, i.e. inflow, finished inflow, historical delayed OD ratio matrix, compared with the prediction of OD matrix at final time slot $t'$.

Despite the contribution of data complete process to our model, we are also interested in the research question that whether or not this data complete estimator can improve the performance of other models? To answer this research question, we do the following experiments. First, we integrate the data completeness process into deep learning baselines and get their variants, i.e. C-ANN, C-FC-LSTM, C-ConvLSTM, C-GCN, C-GEML. Take C-GCN for example, its architecture is shown in Figure \ref{fig:cgcn}. Then we train all the above models and test them to obtain the model performance by repeating each experiments for 10 times and the train/val/test datasets are the same as Table \ref{tab:Results}. The experiment results are shown in Table \ref{tab:var}.

As we can see from both Table \ref{tab:Results} and Table \ref{tab:var}, the data completion process can generally improve the deep learning baselines performance in different extent on different datasets. For example, C-ANN, C-GCN, C-FC-LSTM, C-ConvLSTM, C-GEML improve more 1.1\%, 1.2\%, 1.7\%, 0.3\%, 1.9\% in WMAPE in SZ and 1.2\%, 1.3\%, 1.8\%, 0.5\%, 2.1\% in SH. Note that the improvement in SH is slightly larger than that in SZ, perhaps due to the reason that SH is more incomplete than SZ and SH can benefit more from the data completion process.

\section{Conclusion}
This paper proposes a model C-AHGCSP to conduct short-term OD prediction in rail transit network. The main module of  C-AHGCSP is an Adaptive Heterogeneous Graph Convolution based Spatiotemporal Predictor which extracts multiple heterogeneous spatial correlations (i.e. geographical proximity, region functionality, global dynamicity) with appropriate approaches (i.e. gaussian kernel function, KL divergence, self-attention mechanism). We also propose a data complete estimator to solve the delayed data collection problem in metro scenario, which is composed of a prior estimator and AHGCSP based estimator. Extensive experiments are carried out on two real-world metro datasets and show the superiority of our model over the benchmarks. In addition, our data complete estimator can be also modified to be integrated into other models to improve their model performance.

\bibliographystyle{IEEEtran}
\bibliography{IEEEabrv, ref}

\end{document}